\documentclass[a4, 11pt]{article}
\usepackage{amsfonts, amsthm, amsmath,}
\usepackage{enumitem}
\usepackage[utf8]{inputenc}
\usepackage[english]{babel}
\usepackage{xcolor}
\usepackage{apacite}
\usepackage{color}
\usepackage{nicefrac}
\usepackage{comment} 
\usepackage[hidelinks]{hyperref}
\usepackage[]{natbib}
\usepackage{graphicx}
\usepackage{subcaption}
\usepackage{caption}
\usepackage{authblk}
\usepackage[top=1.8in, bottom=1.5in, left=1.3in, right=1.3in]{geometry}
\begin{document}

	\title{Fairness Hacking: The Malicious Practice of Shrouding Unfairness in Algorithms}

	\author[1,2]{Kristof Meding\thanks{Opinions expressed are solely our own and do not express the views or opinions of our employees or other affiliated parties.}\thanks{kristof.meding@uni-tuebingen.de}}
	\author[3]{Thilo Hagendorff\thanks{thilo.hagendorff@iris.uni-stuttgart.de}}
	\affil[1]{{Data Protection Authority of Baden-Württemberg}, {Germany}}
	\affil[2]{{University of Tübingen}, {Germany}}
	\affil[3]{{University of Stuttgart}, {Germany}}
	
	\date{}
	
	\maketitle
	
	\begin{abstract}
		\noindent Fairness in machine learning (ML) is an ever-growing field of research due to the manifold potential for harm from algorithmic discrimination. To prevent such harm, a large body of literature develops new approaches to quantify fairness. Here, we investigate how one can divert the quantification of fairness by describing a practice we call "fairness hacking" for the purpose of shrouding unfairness in algorithms. This impacts end-users who rely on learning algorithms, as well as the broader community interested in fair AI practices. We introduce two different categories of fairness hacking in reference to the established concept of p-hacking. The first category, intra-metric fairness hacking, describes the misuse of a particular metric by adding or removing sensitive attributes from the analysis. In this context, countermeasures that have been developed to prevent or reduce p-hacking can be applied to similarly prevent or reduce fairness hacking. The second category of fairness hacking is inter-metric fairness hacking. Inter-metric fairness hacking is the search for a specific fair metric with given attributes. We argue that countermeasures to prevent or reduce inter-metric fairness hacking are still in their infancy. Finally, we demonstrate both types of fairness hacking using real datasets. Our paper intends to serve as a guidance for discussions within the fair ML community to prevent or reduce the misuse of fairness metrics, and thus reduce overall harm from ML applications.
		
	\end{abstract}

	\section{Introduction}
Machine learning (ML) algorithms that influence decision-making are being increasingly used within high-stakes fields such as employment, health, policing, trust score assessment, and lending. The trend in ML algorithms used within these fields has rendered the problem of algorithmic discrimination an important emerging issue \citep{https://doi.org/10.48550/arxiv.2010.04053}. Discrimination, legally defined as the unfavorable treatment of individuals due to their belonging to certain demographic groups, is prohibited by international law. In order to establish ethical and trustworthy artificial intelligence (AI) as the field moves forward, it is paramount that new algorithm development prioritizes fairness-aware learning algorithms as well as the avoidance of machine bias \citep{Jobin.2019}. While a vast portion of current publications in the field describes how to define and generate approximations of fairness in AI \citep{Madaio.2020}, an equal proportion of the current literature describes shortcomings in AI fairness performances or shortcomings of the fairness concepts themselves \citep{Weinberg.2022}.

The current study aims to provide a comprehensive and conclusive summary of the shortcomings of current fairness-aware learning algorithms. We describe the adversarial practice of fairness hacking. Fairness hacking describes the adding or removing of (sensitive) attributes from the analysis to lead outsiders to believe that fairness is ensured. Fairness hacking also facilitates the illegitimate exploitation of the plurality of fairness definitions. This occurs when tests are performed using an array of different fairness metrics, but only metrics that produce positive results are reported. During the process of fairness hacking, the fairness shortcomings in ML algorithms are cloaked by reporting that these algorithms have been deemed fair by certain metrics, while any metrics that have produced negative results are suppressed. Fairness hacking often hinges on using the fulfillment of a very specific or intentionally downscaled dimension of fairness to comprehensively report that an algorithm is fair. 

Arguably, the most famous case of fairness hacking unfolded in 2016 when Northpointe—the company that developed COMPAS, a tool for assisting in decisions on whether defendants awaiting trial are too dangerous to be released—defended against allegations from ProPublica. ProPublica published an article claiming that COMPAS is biased against black individuals \citep{Angwin.2016}. COMPAS uses more than 100 factors to calculate a risk factor for defendants, but race is not included in these factors. Nevertheless, \cite{Angwin.2016} claim that for individuals who did not re-offend, black individuals were more likely to receive a medium or high score than white individuals. However, Northpointe claimed that when picking a score between 0 and 10, the calculated chance of recidivism is approximately equal for black and white defendants. In this case, we see the impossibility theorem at work because it is impossible to fulfill Angwin et al.'s as well as Northpointe's fairness criteria at once \citep{CorbettDavies.2016}. Since the overall recidivism rate for black defendants is higher, and the proportion of individuals who re-offend is roughly equal for black and white individuals given an arbitrary risk score, it is inevitable that black people who do not re-offend are assigned a higher risk score than white people who do not re-offend. If COMPAS would fulfill the fairness criteria that Angwin et al. propose, the software could still be viewed as unfair because one would have to systematically assign white individuals higher risk scores than black individuals. This approach is problematic as well. In this instance, we certainly cannot assert that ProPublica or Northpointe intentionally engaged in fairness hacking. However, the ensuing discussion about the appropriate use of fairness metrics perfectly illustrates the problem that at least inter-metric fairness hacking presents.

In this study, we will begin by describing analogies between fairness and p-hacking, since both practices work by selecting and reporting only specific results for data analysis. Next, we will describe intra-metric as well as inter-metric fairness hacking. The former stands for the misuse of particular fairness metrics by adding or removing attributes in the analysis. The latter involves the search for a specific fair metric with given attributes, inverting the impossibility theorem. To conclude the current study, we perform both types of fairness hacking using real datasets. This demonstration provides insight into how fairness hacking may work in real-world scenarios. We hope that this study in its totality will spark discussions on adversarial practices in ML within the fairness, accountability, and transparency (FAT) community. We hope that this will prevent or reduce the misuse of fairness metrics when applied to ML algorithms.

\section{Related work}

Recent work on fairness in ML has demonstrated that fairness targets can be thwarted via technical means. By corrupting data, markedly adversarial attacks against fairness-aware learners can, for example \cite{Jo.2022}, thwart efforts to render AI systems fair. In particular, gradient-based poisoning attacks can cause fairness-relevant classification disparities across specific groups \citep{https://doi.org/10.48550/arxiv.2004.07401}. While this field of research investigates the adversarial introduction of algorithmic discrimination, fairness hacking, in contrast, is about the adversarial introduction of algorithmic fairness. Other adjacent studies have recently compared different fairness metrics without looking at confidence intervals \citep{vzliobaite2017measuring}. This research would benefit from the addition of confidence intervals, as calculating confidence intervals for group-based metrics directly works against fairness hacking. Despite this benefit, very few research studies include confidence or credible intervals \citep{besse2018confidence,ji2020can,ding2021retiring,cherian2023statistical,roy2023fairness}. In addition to the previously mentioned work, other researchers have investigated the performance, as well as techniques used to remove bias, of different datasets and metrics \citep{biswas2020machine}.

 Many different fairness definitions have led theoretical researchers to stipulate various "fairness traps" \citep{Selbst.2018b} or "fairness limitations" \citep{Buyl.2022}. Both fairness traps and fairness limitations can prevent fairness targets from being achieved \citep{Hoffmann.2019, Green.2020}. Fairness traps include the failure to define fairness when applied to a wider sociotechnical system \citep{JohnMathews2023}, the problem of re-purposing algorithmic solutions within different social contexts where these solutions are inappropriate, and the failure to formalize fairness in purely mathematical terms \citep{Weinberg.2022, Castelnovo.2022, Mitchell.2021}. The existence of these fairness traps supports the idea of a non-ideal approach to fairness \citep{Fazelpour.2020}. This non-ideal approach dismisses the idea that the gap between the status quo and conditions of perfect fairness can be closed at all. Metrics for quantifying deviations from fairness are imperfect, and therefore fairness should be framed as a non-ideal methodology that does not include possessing ideal standards and norms. Moreover, one of the critical aspects of fair machine learning is the a priori definition of sensitive categories. As highlighted by \cite{JohnMathews2023}, these categories are often grounded in social constructs and should be carefully selected based on ethical principles and legal guidelines. This aligns with the "realist" approach to fairness, as discussed by \cite{cardon2023displacement}, where fairness is grounded in existing demographic categories produced by institutions. The choice of fairness metrics should not be arbitrary but guided by underlying philosophies of justice. By adopting such a comprehensive approach, investigators can ensure that the machine learning algorithms they develop are not only technically sound but also ethically robust.

Additionally, the impossibility theorem indirectly supports the idea of a non-ideal approach to fairness by showing that no existing method of algorithmic fairness can satisfy different fairness conditions all at once \citep{kleinberg2016inherent, Saravanakumar.2021}. Thus, the impossibility theorem renders fairness into an unachievable ideal. This insight can easily be misused for political reasons. For example, it may be invoked to serve the political motives of those who aim to eliminate efforts to ensure fairness in ML models entirely. However, even for those who seek to ensure fairness within ML models, fairness does not achieve perfect standardization or meet metrics that model perfection \citep{Xivuri.2021, Hanna.2020, Beutel.2019}. Phenomena such as Simpson's paradox \citep{Sharma.2022} and fairness gerrymandering \citep[disadvantaging subgroups despite group fairness,][]{pmlr-v80-kearns18a} further underpin this insight and are similar to fairness hacking. Even in light of these caveats, efforts to increase fairness should still be attempted \citep{Holstein.2019, Lum_2022}. However, measures to increase fairness are always overshadowed by the aforementioned shortcomings. Furthermore, these shortcomings can be harnessed to purport an unfair algorithm to be fair. One way of doing this is through fairness hacking.

\subsection{P-hacking in the sciences}

The term "fairness hacking" as well as its underlying definition were inspired by the concept of p-hacking. Here, we will give a brief introduction to p-hacking before discussing fairness hacking. Essentially, p-hacking is the practice of rendering results that are not statistically significant to appear statistically significant. This is achieved by searching through different combinations of data and statistical approaches until settling on the approach and data input that render the results statistically significant. In the sciences, p-hacking causes an inflation of false positive results \citep{Ioannidis.2005}. This is particularly harmful because false scientific findings circulate. These findings can, in turn, influence other researchers who base follow-up studies on the false positive results of these studies. Researchers have an incentive to manipulate research results because established scientific publication practices mostly reward positive results \citep{fanelli2012negative}. Negative results are seldom published, and it can be challenging to reproduce positive results \citep{open2015estimating}. Moreover, there is strong pressure within the scientific community to produce novel, significant results. For some researchers, this can outweigh the pressure to adhere to sound, rigorous methodology when conducting experiments. These factors coalesce in a way that makes p-hacking tempting for researchers and data analysts of all disciplines. In response, countermeasures to p-hacking have already been developed.

There are two main categories of measures to counteract p-hacking, which will be referred to as non-technical and technical countermeasures. The use of non-technical countermeasures refers to changing entrenched publication practices. One example of a non-technical countermeasure is the practice of registering studies and experiments before performing them with a neutral party \citep{van2016pre,p2021pre}. Another example of a non-technical measure to counteract p-hacking is the publication of negative results \citep{Mehta.2019}. If both of these countermeasure examples were combined, researchers could present their experimental project plan in detail before performing the experiment. Then, the experimental results could be published regardless of the experimental outcome, with both positive and negative results considered worthy of publication. In this example, the pressure on the researcher to publish positive results is reduced. In contrast, technical measures to reduce or prevent p-hacking are directed against specific problems such as the multiple comparison problem \citep{westfall1993resampling}. The multiple comparison problem describes a situation in which variables are added to an analysis until a significant result is found. In this example, any non-significant variables are ignored. In an example of a multiple comparison problem, an analysis is being performed wherein researchers examine the influence of different variables on high blood pressure. For this analysis, the researchers hypothesize that one variable, out of a potentially unlimited number of variables, has an influence on high blood pressure. In this example, an alpha level of 5\% is set for the statistical tests performed as part of this analysis. In this context, a simplified explanation of the alpha level determines how likely it is that a variable is found to influence high blood pressure, although the variable does not truly influence blood pressure. This describes the likelihood of a false positive result. When the analysis is performed, the researchers find that sleep duration has a significant effect on hypertension, but the researchers do not correct the alpha level to account for various multiple comparisons with other variables. When the researchers test an infinite number of measurement variables without accounting for the number of variables, they will certainly find a variable that has a significant influence on blood pressure, even if this variable's influence is not actually true. To prevent such random findings when testing multiple variables, corrections have been proposed \citep{shaffer1995multiple}. A very simple correction is the unweighted Bonferroni correction \citep{bonferroni1936teoria}. The unweighted Bonferroni correction states that the original alpha level in the analysis must be divided by the number (n) of variables tested \(\alpha_{new} = \frac{\alpha}{n}\). Thus, the alpha level is corrected and adjusted according to the number of tested variables. In this review, we will discuss the Bonferroni correction as it applies to fairness hacking (see section \ref{sec::recom}) \footnote{ Please note that the Bonferroni correcting in our examples is used because of the simple implementation and for the sake of the argument. However, there are concerns regarding the reliability \citep{white2019beyond} and methods such as Bayesian statistics or bootstrapping approaches could be used for mitigation \citep{ji2020can,dimitrakakis2019bayesian,foulds2020bayesian,cherian2023statistical}.}.

\subsection{From p-hacking to fairness hacking}
The problem of p-hacking has a long and sad tradition in science. We argue that it is important to build on the work done on p-hacking since fairness hacking is an extended version of p-hacking. In classic p-hacking, researchers are interested in producing statistically significant results. In fairness hacking, however, it depends on the investigator's point of view whether the desired result is the significant or the non-significant one. For example, a company may be interested in a result that is not significant and therefore does not indicate discrimination. Therefore, the goal can be—in contrast to p-hacking—to influence the results in the significant as well as in the non-significant direction.

Furthermore, fairness hacking can occur in at least two variants: intra-metric or inter-metric fairness hacking. Suppose we test the model for discrimination against different attributes (race, sex, age, etc.) within one specific metric. In analogy to medicine and p-hacking, where one would examine the influence of different groups on hypertension by means of a single test (e.g., a t-test), we run into the the multi-comparison problem since we test different groups. We will refer to this type of fairness hacking as intra-metric fairness hacking.

Another big difference in fairness hacking when compared to p-hacking is that there are many different definitions of what fairness entails. Therefore, there are different concepts of how fairness should be measured. This is exploited in inter-metric fairness hacking. A metric is searched until the desired result (discrimination against or not against a group, depending on the desired result) is shown. Then only this single metric, and not all other tested metrics, is reported.

Are intra-metric and inter-metric fairness hacking related? Both practices describe the process of hacking fairness. However, while the former is more related to classical p-hacking, the latter is only possible when multiple metrics are used. Due to the nature of the fairness metrics, they do not necessarily overlap in their outcomes - as implied by the impossibility theorem. Therefore, we argue that both intra- and inter-metric fairness hacking can be used to render an algorithm fair or unfair, even though both practices depict different procedures and involve different methods. 

In the following, we will discuss intra- and inter-metric fairness hacking in detail.

\section{Methods}
Before we start to describe our methods, we would like to clarify our terminology, which is borrowed from \cite{barocas-hardt-narayanan}. With the \textit{features}, we describe (random) variables used as input for a machine learning algorithm. With \textit{(sensitive) attributes}, we describe (random) variables on which the fairness of the outcome of a machine learning algorithm is analyzed. Typical sensitive attributes are race, age, or gender. Within one attribute—such as age—we have different \textit{groups}, such as young and old. 

Our main results are grounded in the use of different fairness metrics. We focus primarily on group-based metrics because we believe that they are the most commonly used in applications. The following group-based metrics, or specifically the difference between the metrics, are used: average odds, base rates, equal opportunity, error rate, false omission, false positives, predictive parity, statistical parity, and true negatives; see \cite{aif360-oct-2018} for definitions. Furthermore, we use the Theil index \citep{speicher2018unified} and consistency \citep{zemel2013learning} as an in-between measure of individual and group-based metrics. 

\noindent First, we use synthetic data for our investigation. In this investigation, we assume that there were 100 participants in our dataset. Every participant has 1,000 randomly Bernoulli (binary) distributed attributes.\footnote{The binary setting here is for illustration purposes only. Our results hold equally for non-binary settings.} These attributes could be, for example, an age variable (young vs. elderly or villager vs. townspeople). In our \emph{theoretical} analysis, the accuracy of our algorithm does not influence the analysis since all attributes share the same accuracy. The accuracy of our hypothetical algorithm is 75\%. One might ask why we used a simulation-based approach for the comparison of group metrics since here one could also calculate analytical solutions. We used a sampling-based approach on purpose because with a sampling approach, we were easily able to compare the group-based metrics to two individual-based metrics.

Individual fairness metrics rely on the assumption that similar individuals should be treated similarly. For the comparison of individuals, therefore, every hypothetical member has one continuous Gaussian distributed (\(\mathcal{N}(\mu=0,\sigma^{2}=1)\)) feature.  

\noindent The calculation of confidence intervals is important. The group-based metrics are essential differences in binomial distribution except for the average odds metrics. For the average odds metric, one averages the  true- and false positive ratio between the groups. It is possible to calculate the confidence intervals of differences for the ratio \(\Delta p\) of two binomial distributions with various methods \citep{newcombe1998two}. We used the Wald method, basically a normal approximation
\begin{flalign}
\Delta p \pm z_{\alpha}\sqrt{\frac{p_1 (1-p_1)}{n_1} \frac{p_2 (1-p_2)}{n_2}}
\end{flalign}
where n denotes the number of samples in group one and group two respectively, and \(z_{\alpha}\) denotes the standard \(1- \frac{\alpha}{2}\) quantile of the standard normal distribution with confidence level \(\alpha\). We check numerically that the Wald confidence interval method yields a good coverage in our setting.

Besides using synthetic data, we use real data from the Medical Expenditure Panel Survey included in the ''AIF360'' package provided by \citep{aif360-oct-2018} to assess fairness issues in machine learning. We include each binary attribute (134 in total) in our analysis. We train a linear logistic regression classifier provided by AIF360 to predict utilization. The training and test split percentages are 90\% to 10\%. Additionally, data and the introductory example from \citep{ding2021retiring} are used in the appendix. 

\noindent We use the python package ''AIF360'' (v. 0.4.0) to compute different metrics and then include  Medical Expenditure Panel Survey  data for our analysis, matplotlib (v. 3.5.2) for plotting, and the statsmodels package (v. 0.13.2) for calculation of confidence intervals. The folktables dataset is used in v. 0.0.11. 

\section{Results}
In the following chapter, we will present our results. First, we discuss intra-metric fairness hacking. We will argue that this is a variant of p-hacking. Second, we discuss inter-metric fairness hacking. Here we will argue that inter-metric fairness hacking is a specific misuse of fairness metrics inherent to these methods. Finally, we show that fairness hacking is possible on real datasets.

\subsection{Intra-metric fairness hacking as a variant of p-hacking}
\begin{figure}[h!]
	\includegraphics[width=0.99\textwidth]{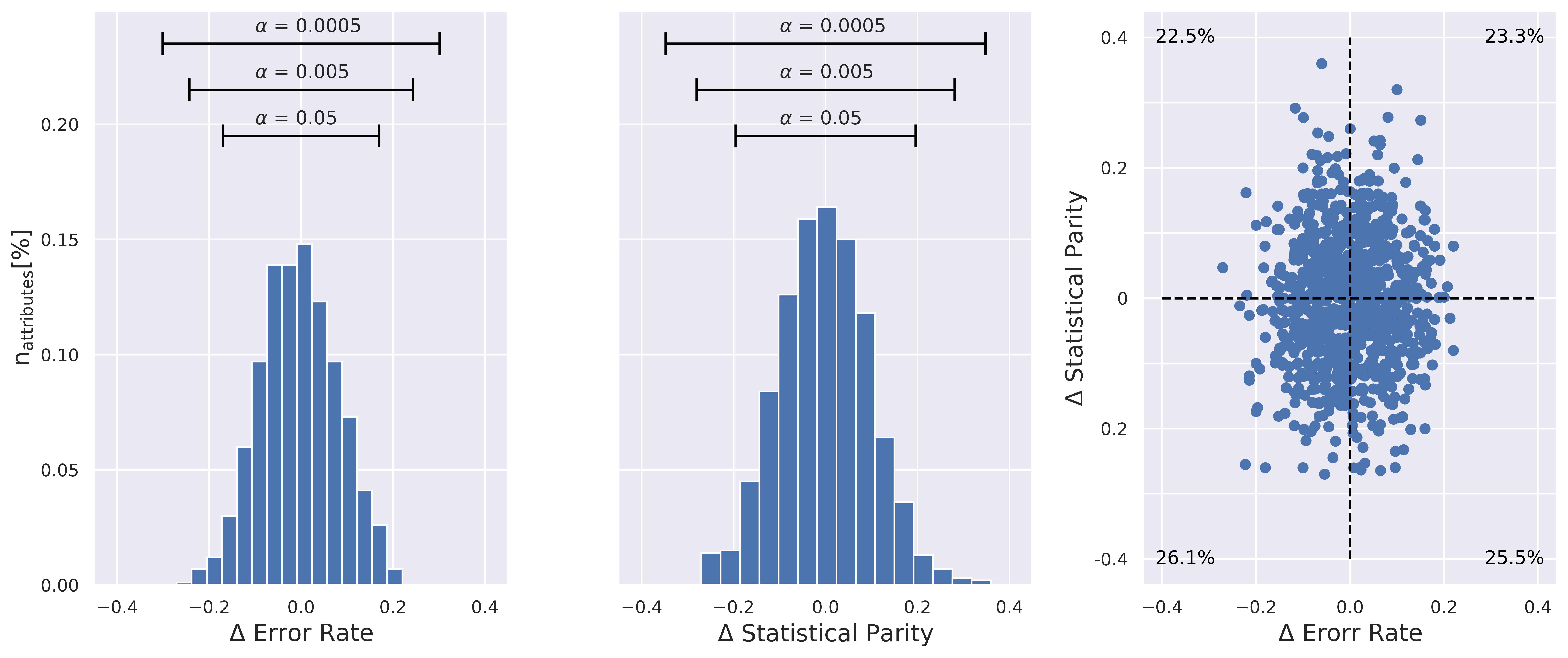}
	\caption{Intra-metric fairness hacking for error rate and statistical parity. Distribution (in percent) of the error rate (left) or statistical parity (center) difference between 1,000 randomly assigned attributes (with binary values) for the outcomes of our hypothetical ML algorithm. A value of zero corresponds to no bias against a group. Values to the left \textit{or} to the right indicate a bias against groups. Upper horizontal lines indicate confidence intervals with different alpha levels. Right: Scatter plot of statistical parity against error rate. The numbers in the corners indicate which percentage of the data falls in the respective quadrant.}
	\label{fig::Figure11}
\end{figure}

Assume that two of our favorite computer scientists, Alice and Bob, participate in a credit scoring study, e.g., similar to the German credit card dataset.\footnote{For more information, please refer to \url{https://archive.ics.uci.edu/ml/datasets/statlog+(german+credit+data).}} Alice and Bob have a number of attributes (age, race, gender, etc.). For simplicity, we argue here that Alice and Bob do not share any group across the attributes—thus, we assume mutually exclusive group membership for all attributes. For the sake of our argument, we next assume an ideal-world scenario. There is no intentional bias on any attribute in the dataset with respect to Alice's and Bob's groups. An algorithm now decides whether a loan is given or not. After one year, the performance of the algorithm is evaluated. Alice and Bob want to investigate whether the algorithm is biased against their respective groups. They agree on a common group-based fairness metric such as equal opportunity. After carefully inspecting the data, both Alice and Bob argue that the algorithm is biased against them for some attributes. How can this happen?

Let us have a look at our data in Figure \ref{fig::Figure11} (left). Here, we plot a histogram of the equal-opportunity difference of the algorithm between all 1,000 attributes. By chance, and inherent to random nature, some group membership identifier will indicate a bias against Alice's or Bob's group. By cherry-picking their favorite group, Alice and Bob can argue that the algorithm is biased against them. Furthermore, an outsider can even argue that the algorithm is not biased since some groups yield zero difference in the metrics. It is important to note that values to the right and to the left both indicate biases of the algorithm since the sign of the value depends on the arbitrary choice of whether the protected group within the attributes is coded zero or one. 

Of course, this implies severe problems. In our case, we introduced by design no bias against attributes \textit{in the synthetic dataset}. Simultaneously, one can find arguments for and against an arbitrary bias. Let us say that this result is obvious. Datasets do always contain biases against certain groups, but maybe only regarding irrelevant attributes. Relevant for fairness discussions are always attributes that include groups that are protected or sensitive. However, we want to highlight the following point: We used 1,000 attributes for illustration purposes. However—and very important—the issue of intra-metric fairness hacking is still present if one uses fewer attributes and/or datasets with a smaller number of participants since smaller datasets tend to have more extreme average values \citep{wainer2007most}. We will discuss inter-metric fairness hacking in real datasets that have attributes (sex, age, race, geographical location, etc.) in section \ref{sec:wild}.

We argue that when intra-metric fairness hacking happens within one fairness metric, it is a form of traditional p-hacking. Group-based fairness metrics are, in essence, a form of signal detection theory within confusion matrices. Thus, intra-metric fairness hacking is a form of the multiple comparisons problem. Luckily, it follows that the designer of our credit score study can use a well-known mechanism of protection against p-hacking to similarly protect against fairness hacking. Here, it is important to highlight that properly adapted confidence intervals, indicated in Figure \ref{fig::Figure11} by the vertical lines, are one method to avoid intra-metric fairness hacking (also see the Discussion section for different approaches).

Does intra-metric fairness hacking depend on the chosen metric? This is not the case. We want to highlight that intra-metric fairness hacking is inherent to fairness metrics. For showing this, we once more plotted Figure \ref{fig::Figure11} with equal opportunity and statistical parity in the appendix; see Figure \ref{fig::Figure12}. Furthermore, intra-metric fairness hacking is not only possible in group-based methods but also in individual fairness metrics—the Theil index\citep{speicher2018unified} and Consistency\citep{zemel2013learning}—as shown in Figure \ref{fig::FigureTheil} and \ref{fig::Consistency} in the appendix. 

\subsection{Inter-metric fairness hacking: inverse of the impossibility theorem}

\begin{figure}[h!]
     \centering
     \begin{subfigure}[b]{.45\textwidth}
         \centering
	\includegraphics[width=1\textwidth]{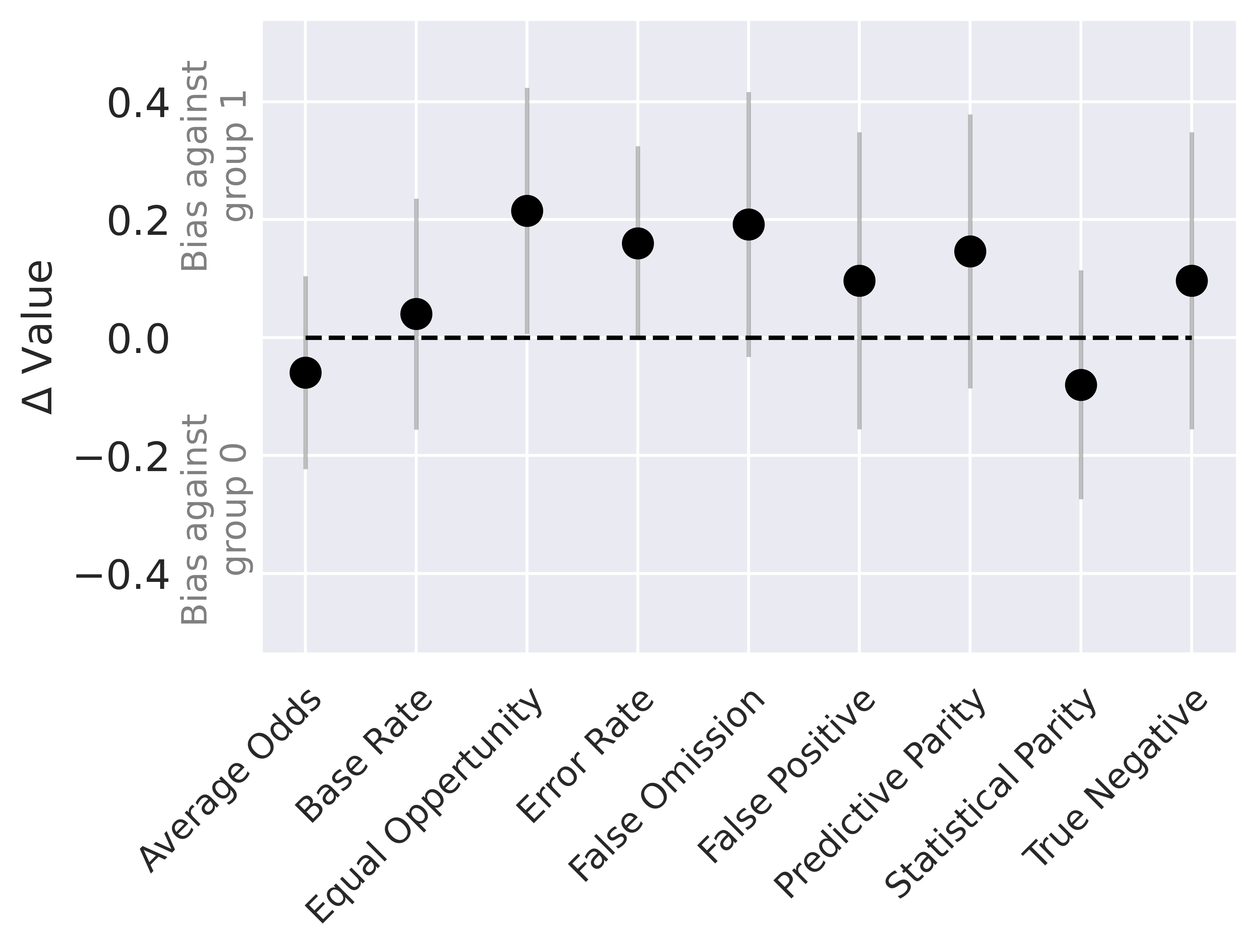}
         \caption{Equal accuracy between groups}
         \label{fig:3_1}
     \end{subfigure}
     \hfill
     \begin{subfigure}[b]{0.45\textwidth}
         \centering
	\includegraphics[width=1\textwidth]{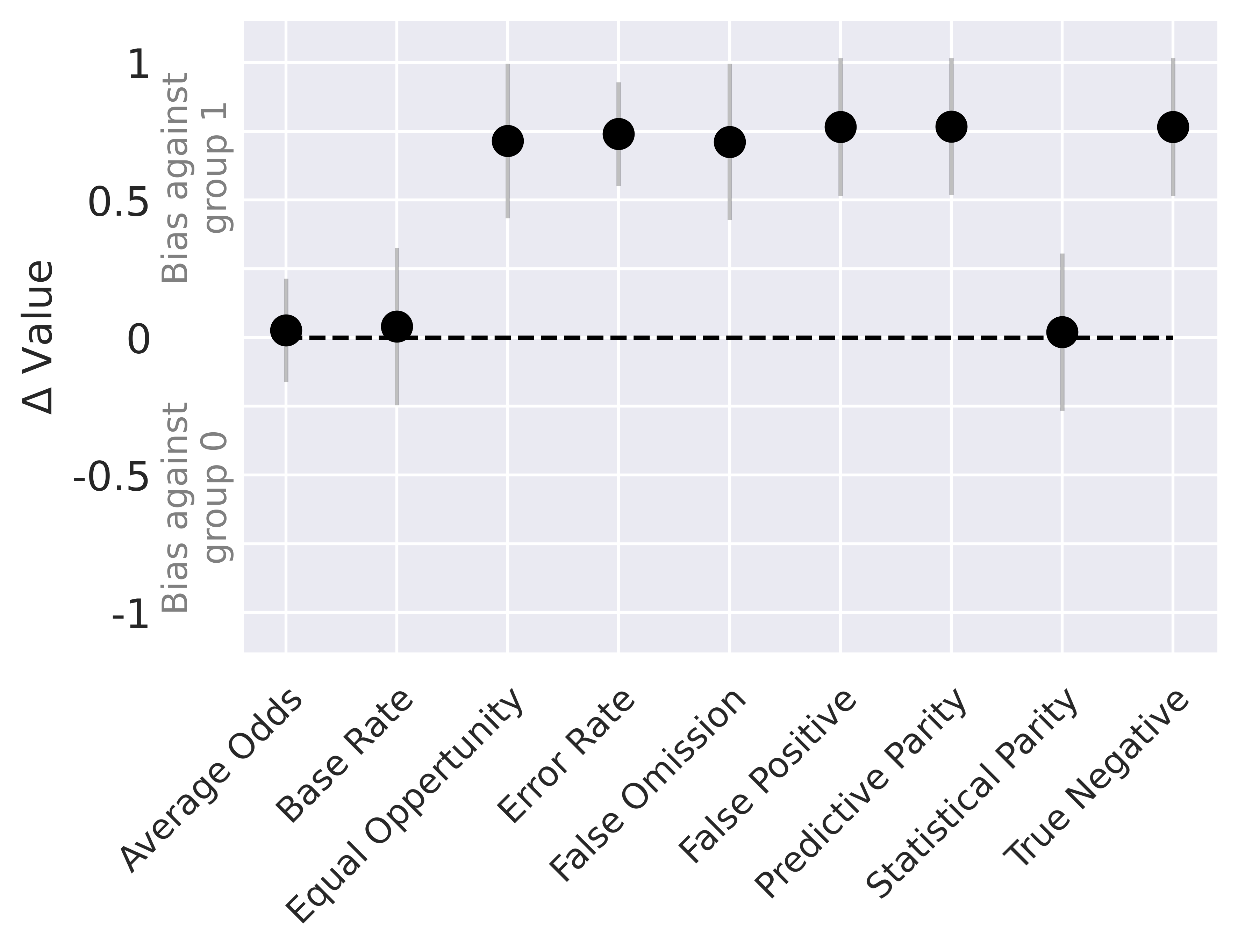}
         \caption{Unequal accuracy between groups}
         \label{fig:3_2}
     \end{subfigure}
        \caption{Inter-metric fairness hacking as the inverse of the impossible fairness theorem. a) Plotted is the difference between group 1 and group 0 of one binary attribute. The binary attribute is randomly distributed across groups. The hypothetical algorithm has 75\% accuracy for both groups. Vertical lines indicate a Bonferroni-corrected confidence interval: alpha level = \(\frac{0.05}{12}\). Figure a) has the same plot conventions as b), but the accuracy difference is set to 0.7.}
        \label{fig:3_ab}
\end{figure}

From now on, we assume that Alice and Bob agree to analyze one single attribute. This single attribute is used to evaluate the fairness of the algorithm. In Figure \ref{fig:3_ab}a, we plot the difference between Alice's (group 0) and Bob's (group 1) groups across a broad range of metrics. It is important to note that plotting fairness metrics without proper confidence intervals would be misleading. For some metrics, group 1 would be seen as unfairly treated; and for other metrics, group 0 would be seen as the one that is unfairly treated. Plotting confidence intervals shows that there is no significant effect for differences across metrics. 

How does the situation change if we intentionally insert a bias in our dataset? To highlight our argument, we inserted a very strong accuracy difference in our algorithm. For Figure \ref{fig:3_ab}a, we assumed an accuracy difference of 0.7 between group 0 and group 1 for our hypothetical algorithm. Here, we now see that several algorithms show a significant bias against group 1. However, and that is very important to note, three metrics do not indicate a bias: average odds, base rates, and statistical parity. Depending on the metrics, Alice and Bob can argue for or against discrimination in group 1. Please note that this effect is not an effect of classic p-hacking since there is a significant effect caused by the accuracy difference. We also used Bonferroni-corrected confidence intervals. Here, choosing the metric instead of the sensitive attribute makes the difference between inter-metric and intra-metric fairness hacking. 

Inter-metric fairness hacking has a relation to the well-known impossibility theorem \citep{kleinberg2016inherent}. The impossibility theorem  claims that it is not possible to satisfy the three metrics of demographic parity, equalized odds, and (positive/negative) predictive parity at the same time. The theorem focuses on the consensus of fairness metrics. However, we have a different focus. We focus on the non-consensus of fairness metrics. The impossibility theorem also implies that at most, two of the three metrics can be fulfilled simultaneously. Therefore, at least one metric does not agree with the other two metrics. This disagreement of metrics is what we define in a more general approach—beyond demographic parity, equalized odds, and positive/negative predictive parity—as inter-metric fairness hacking.

Inter-metric fairness hacking is different from classic p-hacking. It is also different from the problem of choosing the right test statistic in classic hypothesis testing. The correct test statistic depends on the data distribution and assumptions about it. In contrast, from a purely statistical viewpoint, using demographic parity or predictive parity is equally valid.

Next, we demonstrate fairness hacking scenarios using real data.

\subsection{Fairness hacking in real-world datasets}
\label{sec:wild}
\begin{figure}[h!]
     \centering
          \begin{subfigure}[b]{1\textwidth}
         \centering
         \includegraphics[width=1\textwidth]{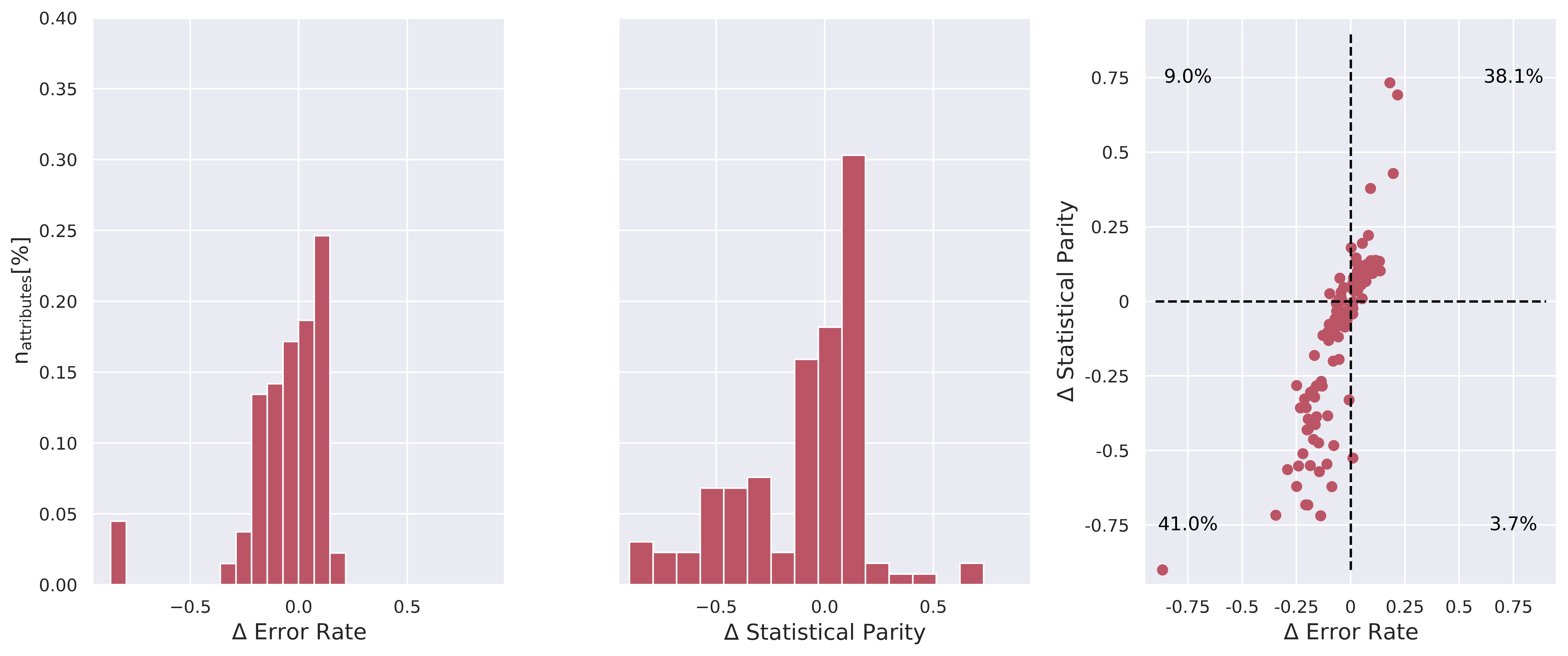}
         \caption{Error rate and statistical parity for all 134 binary attributes. Plot conventions as in Figure a.}
         \label{fig:4_2}
     \end{subfigure}
     \hfill
     \begin{subfigure}[b]{0.45\textwidth}
         \centering
	\includegraphics[width=1\textwidth]{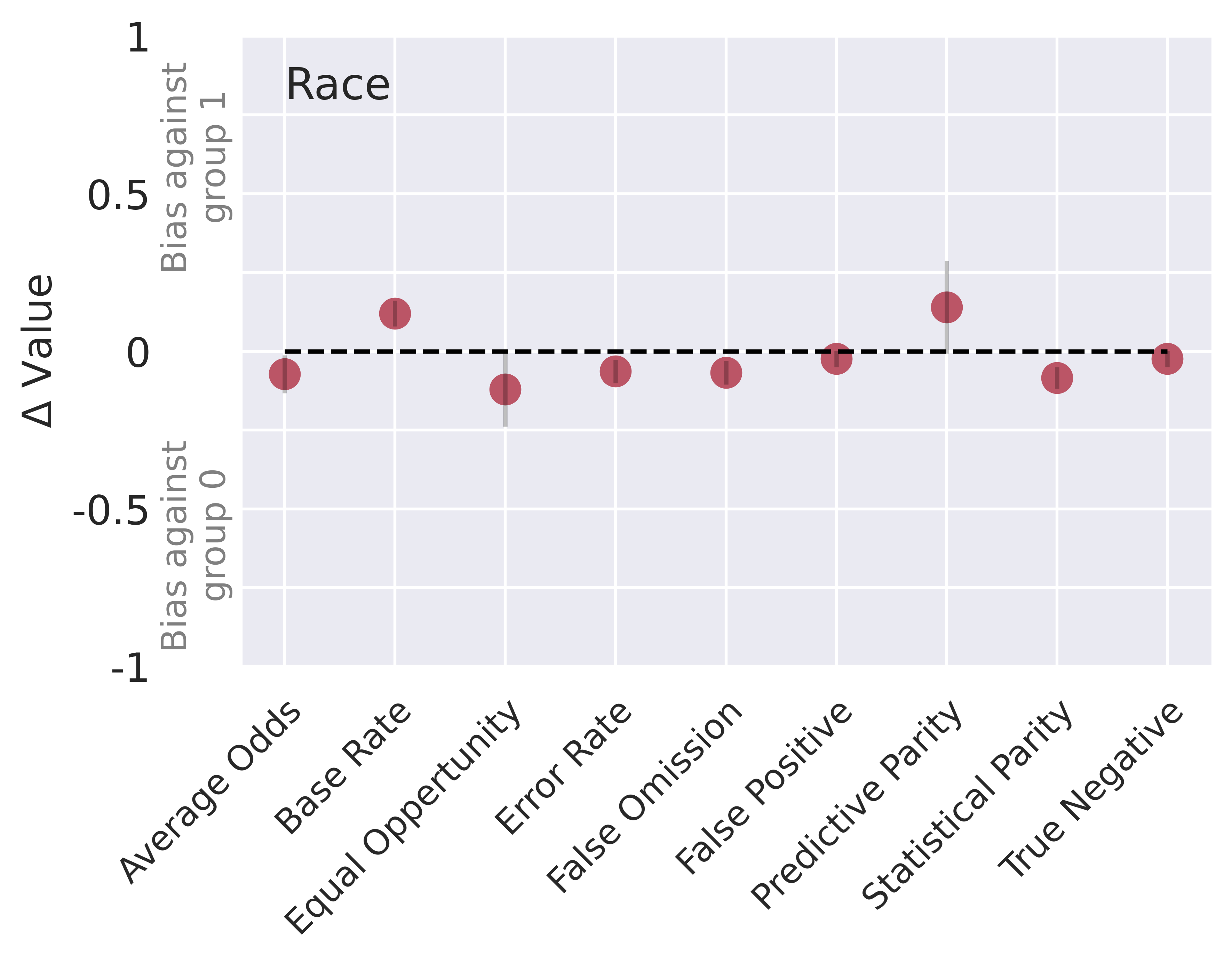}
         \caption{Different fairness metrics for the race attributes. Plot conventions as in Figure 3.}
         \label{fig:4_3}
     \end{subfigure}
     \hfill
      \begin{subfigure}[b]{0.45\textwidth}
         \centering
	\includegraphics[width=1\textwidth]{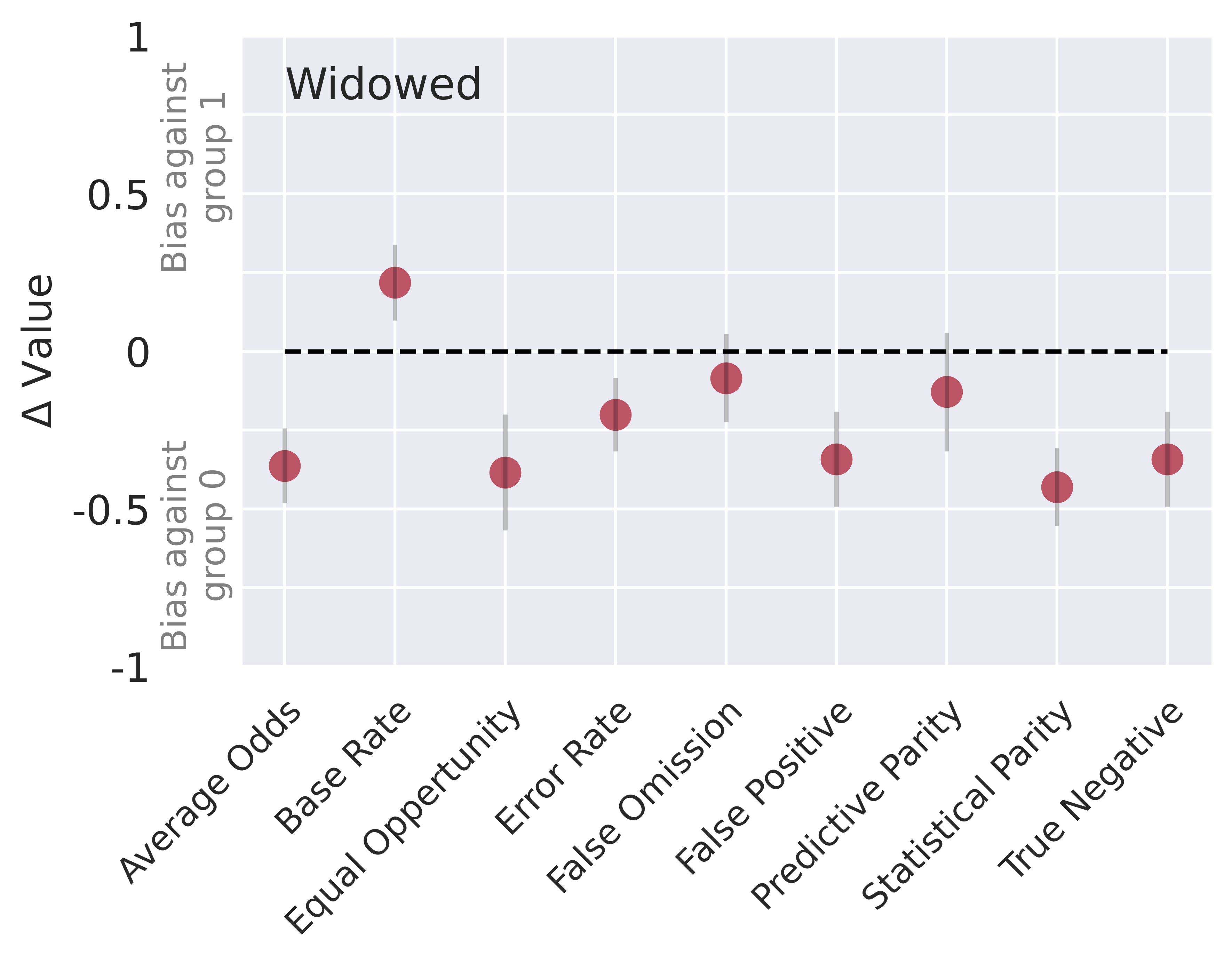}
  \caption{Different fairness metrics for the widowed attributes. Plot conventions as in Figure 3.}
         \label{fig:4_4}
     \end{subfigure}
        \caption{Fairness hacking in the wild for the Medical Expenditure Panel Survey data.}
        \label{fig:4_abc}
\end{figure}
We demonstrate intra- and inter-metric fairness hacking using real data in Figure \ref{fig:4_abc}. Here we used the Medical Expenditure Panel Survey (MEPS) dataset. The goal is to forecast the utilization of patients based on different attributes. We used 138 features to predict utilization, 134 of which are binary attributes used in the fairness analysis. Accuracy on the test split was 86.4\%. Figure \ref{fig:4_2} shows the distribution of the difference between the error rate and statistical parity for all 134 binary attributes. For a comparison of the equal opportunity metric and statistical parity, see Figure \ref{fig:4_1} in the appendix.

We will use the race attribute from the MEPS dataset for intra-metric fairness hacking. Table \ref{tab:my-table} shows the result for corrected and uncorrected confidence intervals for all three metrics (statistical parity, equal opportunity, and error rate) from Figure \ref{fig:4_2}. The adjusted confidence intervals are calculated with the Bonferroni correction. Intra-metric fairness hacking could occur for the first two metrics since the corrected confidence intervals include the zero-value and thus a significant result change toward a non-significant result. This indicates the importance of adjusting confidence intervals. For the error rate metric, uncorrected and corrected confidence intervals yield significant results.

\begin{table}[]
\begin{tabular}{lllll}
                  & Avg. \(\Delta\) & Unc. alpha level & Corrected alpha level &  \\
Equal Opportunity & -12.1\%    & {[}-23.8\%,-0.3\%{]}  & \textbf{{[}-33.4\%,9.3\%{]}}      &  \\
Error Rate        & -6.4\%     & {[}-10.1\%,-2.7\%{]}   & \textbf{{[}-13.1\%,0.4\%{]}}               & \\
Statistical Parity & -8.4\%     & {[}-11.8\%,-5.0\%{]}  & {{[}-14.6\%,-2.2\%{]}}      &  
\end{tabular}
\centering
\caption{Intra-metric fairness hacking for the race attribute in the MEPS dataset. Comparison of the uncorrected \(\alpha=0.05\) and corrected \(\alpha_{corr} = \frac{0.05}{134}\) levels. Bold font indicates a metric changing from significant to non-significant results (including zero) by using the Bonferroni correction.}
\label{tab:my-table}
\end{table}

In addition to intra-metric fairness hacking, inter-metric fairness hacking is likewise a serious risk. We show inter-metric fairness hacking for two attributes (race and widowed) in Figure  \ref{fig:4_3} and Figure \ref{fig:4_4}. When we look at the race attribute, we see that 4 out of 12 metrics indicate a bias against white people, while the remaining 8 metrics indicate a bias against black people. For the widowed attribute, we see that 2 out of 12  metrics indicate a bias against widowed people while the remaining metrics indicate a bias against the group of non-widowed people. A malicious machine learning engineer could use a selected metric to argue for a bias against one or the other group, depending on the chosen metric while neglecting all other metrics.

Taken together, both inter- and intra-metric fairness hacking pose a risk during the development and monitoring of machine learning models.

\section{Recommendations to avoid fairness hacking}
\label{sec::recom}
Fairness hacking can be (possibly) avoided to a certain degree. Hence, having pointed out the problems, we also suggest potential solutions for fairness hacking, both in its intra- as well as inter-metric form. We would like to differentiate between measures that can be implemented easily and immediately as well as measures that require time and concerted effort. We will borrow these ideas and measures developed in the context of p-hacking here, see insofar \cite{head2015extent,stefan2023big} as a good overview. Although complete guidelines on how to avoid fairness hacking are beyond the scope of our paper, we would like to give recommendations. These recommendations are not intended to be generally binding for the future, but rather to form guidelines from our point of view today.

\subsection{Immediate Recommendations}
As a malicious practice, fairness hacking cannot be aligned with the claims of good scientific practice. However, recommendations stemming from the field of p-hacking avoidance can help to avoid inter- and intra-metric fairness hacking as well.

Based on the results presented above, we recommend to always calculate uncertainty intervals. Reporting single values without reporting the significance of the result can be misleading. Choosing p-values for confidence intervals then needs critical thinking in advance \citep{wasserstein2016asa}. \\
However, just because results are significant, this does not necessarily mean that the results have an impact on the real world, too. There can be small effects that are highly significant \citep{cumming2014new,sullivan2012using}. For example, measuring the height of two groups of adults can give a highly statistical difference of 1 mm. However, this effect may have negligible consequences in practice. To prevent this style of reporting statistically significant but otherwise hardly relevant results, the concept of effect size has been developed \citep{cohen1969statistical,sullivan2012using}. Only effects with a sufficient magnitude should be considered statistically sound. This also applies for fairness metrics in the context of significance testing. In machine learning studies, only significant fairness effects with sufficient effect size should be considered.

Intra-metric fairness hacking is kind of similar to p-hacking. Hence, excluding variables in post-analysis is problematic \cite{nosek2012scientific, john2012measuring}, as well as building the hypothesis after the results are known (also called Harking, \cite{kerr1998harking}). These practices are problematic due to them allowing to find significant results through hypothesis testing easily. Similarly, in fairness hacking, some individuals might have an interest in finding a significant discrimination while others might have the opposite interest. To avoid these pitfalls, we recommend two methods based on the p-hacking literature \citep{head2015extent,Ioannidis.2005}. First, one builds a clear hypothesis of which fairness attributes should be included in the analysis \emph{before} one analyses the data. This becomes even more important since the sensitive attributes are grounded on social categories \citep{john2022crisis}. Here, defining a priori sensitive categories based on hypothesis, principles, and regulations before applying fairness metrics is a preventive measure for fairness hacking. In this context, it is also imperative to recognize that sensitive categories are not mere attributes available for arbitrary selection, but are deeply rooted in social constructs. Traditional statistical methods, which have been foundational in shaping our understanding of the social world, are now being re-evaluated in the face of evolving AI models that seek to capture the intricate nuances of social life. This involves grounding the definition of sensitive categories in established ethical principles and sociological theories. In addition to defining sensitive categories a priori, one can label studies as explanatory research \citep{tukey1977exploratory,cumming2014new} ---which is valuable on its own--- without a predefined hypothesis. Here, researchers might also include variables after collecting data. By labeling studies as explanatory research, recipients can treat the result with caution \citep{head2015extent}. Furthermore, when testing multiple variables ---be it hypothesis-driven or in exploratory research---, confidence intervals need to be adapted accordingly. When not following these recommendations, it becomes difficult to detect intra-metric fairness hacking since one simply cannot know the study's confidence and whether multiple testing issues were considered.

Inter-metric fairness hacking shares challenges that another type of studies, namely meta-analyses, are facing. In meta-analyses, separate studies with different methods are compared and summarized, see \citep{borenstein2021introduction}. To avoid problematic outcomes due to the different statistical methods used in different studies, it was recommended to consider potential failures of test designs \citep{carter2019correcting}\footnote{For example, if studies use t-tests one should check when these methods fail (such as small non-normal distributed samples, \cite{lumley2002importance})}. Similarly, in inter-metric fairness hacking scenarios, different statistical tests need to be compared. It is highly important to think in advance about which metric is appropriate to apply, e.g. practitioners can choose a metric based on a theory of justice \citep{cardon2023displacement}. The decision about which metric to use needs to be adequately justified. Contrary to intra-metric fairness hacking, exploratory research may not hold as much value in terms of inter-metric fairness hacking. It may not be appropriate to report every fairness metrics as different metrics suggest distinct perspectives on fairness, which can lead to contradictions. Only methodologically justified metrics should be taken into consideration and reported. Conversely, one could argue that individuals who are affected by algorithmic decisions should be given priority, favoring participatory approaches. If all fairness metrics are displayed, it must be clarified why the results of some metrics have been dismissed. Besides, presenting all metrics aids in uncovering hidden morally relevant assumptions.  

All the above-described recommendations can in principle be applied immediately.

\subsection{Recommendations in the long run}

We would like to discuss four future directions to prevent fairness hacking which are again largely borrowed from the statistics literature: pre-registration, new datasets, checklists, and building good scientific practices.

A way to prevent p-hacking in the context of studies is pre-registration \citep{wagenmakers2012agenda,chambers2013registered}. Here, studies and methods are defined and pre-registered in journals before collecting and analyzing data. This ensures that researchers have to deal with potential problems with their methodology, experimental designs, or metrics before constructing, training, and testing a model. The process could then include a discussion about which metrics should be used, including their advantages and disadvantages. Here, \cite{gundersen2021case} discusses pre-registered reports in the context of ML research which could be adapted to fairness issues.

In addition to registered reports, new datasets or methods can also help mitigate the issues. In our synthetic data, we have shown that fairness hacking can occur from purely statistical noise alone. Through the MEPS data, we show that fairness hacking also occurs in real data, which may have an imbalance in accuracy between groups. Nevertheless, we think that approaches such as those from \cite{ding2021retiring}, which provide a balanced dataset of the known UCI adult dataset, are a reasonable approach. As in the example, we show in the appendix that intra-fairness hacking possibilities are reduced, see Figure\ref{fig::FigureFolkTables}.

In the context of p-hacking, checklists and guidelines for the prevention have been presented \cite{wicherts2016degrees}. These guidelines should help researchers to avoid p-hacking and facilitate good statically reporting. In the context of machine learning, among others, datasheets for datasets \citep{gebru2021datasheets} and model cards for model reporting \citep{mitchell2019model} have been introduced. These checklists focus on the whole life-cycle of machine learning models and their datasets. Also, checklists for fairness were already developed \citep{madaio2020co,agarwal2023seven}. We want to stress that these checklists and guidelines are a first good step in the right direction. However, we hope that the machine learning community will bring up further checklists, especially with regard to the problem of inter- and intra-metric fairness hacking. For this purpose, the calculation of confidence intervals and comparing metrics should, next to becoming good scientific practice, be included in the software packages for the calculation of metrics. Furthermore and so far, we know that none of the standard fairness packages have implementations of confidence intervals by default.

For both types of fairness hacking discussed in this paper, the awareness in the machine learning community has to be raised. Both AI ethics and technical fairness research should include reflections on the potential dangers of fairness hacking into their considerations \citep{Mehrabi.2019}. Often, journals do require solid statistical tests to prevent p-hacking. Otherwise, papers are desk-rejected. This procedure is missing with respect to fairness metrics, although they are statistical tests, too. Here it might make sense that journals need to take fairness issues more into account to prevent the practices described in this study.

In summary, care must be taken during the full life-cycle of development, application and reporting of a machine learning algorithm to justify the choice of a metric whose aptitude for a specific purpose should be critically reflected using not just moral intuitions, but at best sound ethical considerations that take into account all affected individuals. In addition to that, the proprietary nature of many algorithms makes it difficult to detect, let alone rectify, instances of fairness hacking. When algorithms are closed off from public scrutiny, they may mask inherent biases or intentional manipulations that distort the true fairness of the system. Furthermore, the incentives driving fairness hacking are distinct from those behind p-hacking. While p-hacking arises from the pressure to produce statistically significant results, often leading researchers to manipulate or cherry-pick data to achieve this, fairness hacking may stem from commercial or social pressures to showcase unbiased AI models, prompting developers to superficially adjust outputs without addressing underlying biases.

\section{Summary and outlook}

Bias reduction efforts require a mix of technical and non-technical solutions. At least when looking at technical solutions, interventions to increase fairness tend to cluster into associated modes, meaning that different fairness preservation measures strongly correlate with one another {\citep{friedler2018comparative}}. Nevertheless, our paper demonstrates that fairness interventions can be brittle when engaging in intentional fairness hacking. In this context, one has to stress that several phenomena resemble fairness hacking practices, but are not explicitly called fairness hacking. These range from inversion of the impossibility theorem to phenomena such as Simpson's paradox or fairness gerrymandering. For utility-based approaches, one must also keep in mind that the choice of approach depends on the specific problem and the desired trade-off between accuracy and fairness, although this trade-off cannot be considered fairness hacking in the sense proposed in our paper. Ultimately, actual fairness hacking causes a twofold burden. If one picks a single fairness metric and claims that a particular algorithm is unfair without shedding light on the outcomes of alternative metrics, a distorted image of an algorithm's performance is created. On the other hand, if one lets several metrics compete with each other, fairness issues become relative. Both burdens should be avoided. To do this, one must always consider fairness metrics in a given sociotechnical context. One must consider when human judgment and moral intuitions are necessary when weighing different metrics against each other in a given situation. In fact, each fairness metric has an underlying moral assumption that can be explicitly defined \citep{https://doi.org/10.48550/arxiv.1809.03400}. Whereas the term "impossibility theorem" suggest that fairness metrics have the same status and can hence be played off against each other—a circumstance that is inversely exploited by fairness hacking practices—this does not hold on closer examination and when taking into account the normative presuppositions embedded in each fairness metric as well as the suitability of particular fairness metrics in a given context \citep{https://doi.org/10.48550/arxiv.2102.08453, https://doi.org/10.48550/arxiv.1609.07236, https://doi.org/10.48550/arxiv.1710.03184}. Thus, our results do not just underpin the importance of being aware of the ease of claiming a machine learning model to be fair; they also stress the importance of reflecting on the normative values embedded in fairness metrics.

\section*{Acknowledgements}
We would like to thank Frank Jäkel and Moritz Hardt for valuable feedback on an earlier manuscript and Larissa Hoefling for valuable discussions. Funding was provided by  Deutsche Forschungsgemeinschaft (DFG; the German Research Foundation)—project number 276693517—SFB 1233, TP 4 Causal inference strategies in human vision for KM and his work at the University of Tübingen. TH was supported by the Baden-Württemberg  Ministry of Science, Research, and the Arts under Az. 33-7533-9-19/54/5 in Reflecting Intelligent Systems for Diversity, Demography and Democracy (IRIS3D) as well as the Interchange Forum for Reflecting on Intelligent Systems (IRIS) at the University of Stuttgart.

\section*{Author contributions}
The project was initialized by KM during his work at the University of Tübingen and later jointly developed forward with TH during KM's work at the data protection authority of Baden-Württemberg. KM wrote the code for the analysis with input from TH. Both authors planned, structured, and wrote the manuscript.

\bibliography{sample-base}

\begin{thebibliography}{}

\bibitem [\protect \citeauthoryear {%
Agarwal%
\ \BBA {} Agarwal%
}{%
Agarwal%
\ \BBA {} Agarwal%
}{%
{\protect \APACyear {2023}}%
}]{%
agarwal2023seven}
\APACinsertmetastar {%
agarwal2023seven}%
\begin{APACrefauthors}%
Agarwal, A.%
\BCBT {}\ \BBA {} Agarwal, H.%
\end{APACrefauthors}%
\unskip\
\newblock
\APACrefYearMonthDay{2023}{}{}.
\newblock
{\BBOQ}\APACrefatitle {A seven-layer model with checklists for standardising
  fairness assessment throughout the AI lifecycle} {A seven-layer model with
  checklists for standardising fairness assessment throughout the ai
  lifecycle}.{\BBCQ}
\newblock
\APACjournalVolNumPages{AI and Ethics}{}{}{1--16}.
\PrintBackRefs{\CurrentBib}

\bibitem [\protect \citeauthoryear {%
Angwin%
, Larson%
, Mattu%
\BCBL {}\ \BBA {} Kirchner%
}{%
Angwin%
\ \protect \BOthers {.}}{%
{\protect \APACyear {2016}}%
}]{%
Angwin.2016}
\APACinsertmetastar {%
Angwin.2016}%
\begin{APACrefauthors}%
Angwin, J.%
, Larson, J.%
, Mattu, S.%
\BCBL {}\ \BBA {} Kirchner, L.%
\end{APACrefauthors}%
\unskip\
\newblock
\APACrefYearMonthDay{2016}{}{}.
\newblock
\APACrefbtitle {Machine Bias: There's software used across the country to
  predict future criminals. And it's biased against blacks.} {Machine bias:
  There's software used across the country to predict future criminals. and
  it's biased against blacks.}
\newblock
\begin{APACrefURL}
  [{18.01.2018}]\url{https://www.propublica.org/article/machine-bias-risk-assessments-in-criminal-sentencing}
  \end{APACrefURL}
\PrintBackRefs{\CurrentBib}

\bibitem [\protect \citeauthoryear {%
Barocas%
, Hardt%
\BCBL {}\ \BBA {} Narayanan%
}{%
Barocas%
\ \protect \BOthers {.}}{%
{\protect \APACyear {2019}}%
}]{%
barocas-hardt-narayanan}
\APACinsertmetastar {%
barocas-hardt-narayanan}%
\begin{APACrefauthors}%
Barocas, S.%
, Hardt, M.%
\BCBL {}\ \BBA {} Narayanan, A.%
\end{APACrefauthors}%
\unskip\
\newblock
\APACrefYear{2019}.
\newblock
\APACrefbtitle {Fairness and Machine Learning: Limitations and Opportunities}
  {Fairness and machine learning: Limitations and opportunities}.
\newblock
\APACaddressPublisher{}{fairmlbook.org}.
\newblock
\APACrefnote{\url{http://www.fairmlbook.org}}
\PrintBackRefs{\CurrentBib}

\bibitem [\protect \citeauthoryear {%
Bellamy%
\ \protect \BOthers {.}}{%
Bellamy%
\ \protect \BOthers {.}}{%
{\protect \APACyear {2018}}%
}]{%
aif360-oct-2018}
\APACinsertmetastar {%
aif360-oct-2018}%
\begin{APACrefauthors}%
Bellamy, R\BPBI K\BPBI E.%
, Dey, K.%
, Hind, M.%
, Hoffman, S\BPBI C.%
, Houde, S.%
, Kannan, K.%
\BDBL {}Zhang, Y.%
\end{APACrefauthors}%
\unskip\
\newblock
\APACrefYearMonthDay{2018}{{\APACmonth{10}}}{}.
\newblock
\APACrefbtitle {{AI Fairness} 360: An Extensible Toolkit for Detecting,
  Understanding, and Mitigating Unwanted Algorithmic Bias.} {{AI Fairness} 360:
  An extensible toolkit for detecting, understanding, and mitigating unwanted
  algorithmic bias.}
\PrintBackRefs{\CurrentBib}

\bibitem [\protect \citeauthoryear {%
Besse%
, del Barrio%
, Gordaliza%
\BCBL {}\ \BBA {} Loubes%
}{%
Besse%
\ \protect \BOthers {.}}{%
{\protect \APACyear {2018}}%
}]{%
besse2018confidence}
\APACinsertmetastar {%
besse2018confidence}%
\begin{APACrefauthors}%
Besse, P.%
, del Barrio, E.%
, Gordaliza, P.%
\BCBL {}\ \BBA {} Loubes, J\BHBI M.%
\end{APACrefauthors}%
\unskip\
\newblock
\APACrefYearMonthDay{2018}{}{}.
\newblock
{\BBOQ}\APACrefatitle {Confidence intervals for testing disparate impact in
  fair learning} {Confidence intervals for testing disparate impact in fair
  learning}.{\BBCQ}
\newblock
\APACjournalVolNumPages{arXiv preprint arXiv:1807.06362}{}{}{}.
\PrintBackRefs{\CurrentBib}

\bibitem [\protect \citeauthoryear {%
Beutel%
\ \protect \BOthers {.}}{%
Beutel%
\ \protect \BOthers {.}}{%
{\protect \APACyear {2019}}%
}]{%
Beutel.2019}
\APACinsertmetastar {%
Beutel.2019}%
\begin{APACrefauthors}%
Beutel, A.%
, Chen, J.%
, Doshi, T.%
, Qian, H.%
, Woodruff, A.%
, Luu, C.%
\BDBL {}Chi, E\BPBI H.%
\end{APACrefauthors}%
\unskip\
\newblock
\APACrefYearMonthDay{2019}{}{}.
\newblock
{\BBOQ}\APACrefatitle {Putting Fairness Principles into Practice: Challenges,
  Metrics, and Improvements} {Putting fairness principles into practice:
  Challenges, metrics, and improvements}.{\BBCQ}
\newblock
\APACjournalVolNumPages{arXiv preprint arXiv:1901.04562}{}{}{}.
\PrintBackRefs{\CurrentBib}

\bibitem [\protect \citeauthoryear {%
Biswas%
\ \BBA {} Rajan%
}{%
Biswas%
\ \BBA {} Rajan%
}{%
{\protect \APACyear {2020}}%
}]{%
biswas2020machine}
\APACinsertmetastar {%
biswas2020machine}%
\begin{APACrefauthors}%
Biswas, S.%
\BCBT {}\ \BBA {} Rajan, H.%
\end{APACrefauthors}%
\unskip\
\newblock
\APACrefYearMonthDay{2020}{}{}.
\newblock
{\BBOQ}\APACrefatitle {Do the machine learning models on a crowd sourced
  platform exhibit bias? an empirical study on model fairness} {Do the machine
  learning models on a crowd sourced platform exhibit bias? an empirical study
  on model fairness}.{\BBCQ}
\newblock
\BIn{} \APACrefbtitle {Proceedings of the 28th ACM joint meeting on European
  software engineering conference and symposium on the foundations of software
  engineering} {Proceedings of the 28th acm joint meeting on european software
  engineering conference and symposium on the foundations of software
  engineering}\ (\BPGS\ 642--653).
\PrintBackRefs{\CurrentBib}

\bibitem [\protect \citeauthoryear {%
Bonferroni%
}{%
Bonferroni%
}{%
{\protect \APACyear {1936}}%
}]{%
bonferroni1936teoria}
\APACinsertmetastar {%
bonferroni1936teoria}%
\begin{APACrefauthors}%
Bonferroni, C.%
\end{APACrefauthors}%
\unskip\
\newblock
\APACrefYearMonthDay{1936}{}{}.
\newblock
{\BBOQ}\APACrefatitle {Teoria statistica delle classi e calcolo delle
  probabilita} {Teoria statistica delle classi e calcolo delle
  probabilita}.{\BBCQ}
\newblock
\APACjournalVolNumPages{Pubblicazioni del R Istituto Superiore di Scienze
  Economiche e Commericiali di Firenze}{8}{}{3--62}.
\PrintBackRefs{\CurrentBib}

\bibitem [\protect \citeauthoryear {%
Borenstein%
, Hedges%
, Higgins%
\BCBL {}\ \BBA {} Rothstein%
}{%
Borenstein%
\ \protect \BOthers {.}}{%
{\protect \APACyear {2009}}%
}]{%
borenstein2021introduction}
\APACinsertmetastar {%
borenstein2021introduction}%
\begin{APACrefauthors}%
Borenstein, M.%
, Hedges, L\BPBI V.%
, Higgins, J\BPBI P.%
\BCBL {}\ \BBA {} Rothstein, H\BPBI R.%
\end{APACrefauthors}%
\unskip\
\newblock
\APACrefYear{2009}.
\newblock
\APACrefbtitle {Introduction to meta-analysis} {Introduction to meta-analysis}.
\newblock
\APACaddressPublisher{}{John Wiley \& Sons}.
\PrintBackRefs{\CurrentBib}

\bibitem [\protect \citeauthoryear {%
Buyl%
\ \BBA {} de Bie%
}{%
Buyl%
\ \BBA {} de Bie%
}{%
{\protect \APACyear {2022}}%
}]{%
Buyl.2022}
\APACinsertmetastar {%
Buyl.2022}%
\begin{APACrefauthors}%
Buyl, M.%
\BCBT {}\ \BBA {} de Bie, T.%
\end{APACrefauthors}%
\unskip\
\newblock
\APACrefYearMonthDay{2022}{}{}.
\newblock
{\BBOQ}\APACrefatitle {Inherent Limitations of AI Fairness} {Inherent
  limitations of ai fairness}.{\BBCQ}
\newblock
\APACjournalVolNumPages{arXiv preprint arXiv:2212.06495}{}{}{}.
\PrintBackRefs{\CurrentBib}

\bibitem [\protect \citeauthoryear {%
Cardon%
\ \BBA {} John-Mathews%
}{%
Cardon%
\ \BBA {} John-Mathews%
}{%
{\protect \APACyear {2023}}%
}]{%
cardon2023displacement}
\APACinsertmetastar {%
cardon2023displacement}%
\begin{APACrefauthors}%
Cardon, D.%
\BCBT {}\ \BBA {} John-Mathews, J\BHBI M.%
\end{APACrefauthors}%
\unskip\
\newblock
\APACrefYearMonthDay{2023}{}{}.
\newblock
{\BBOQ}\APACrefatitle {The displacement of reality tests. The selection of
  individuals in the age of machine learning} {The displacement of reality
  tests. the selection of individuals in the age of machine learning}.{\BBCQ}
\newblock
\APACjournalVolNumPages{Distinktion: Journal of Social Theory}{}{}{1--24}.
\PrintBackRefs{\CurrentBib}

\bibitem [\protect \citeauthoryear {%
Carter%
, Sch{\"o}nbrodt%
, Gervais%
\BCBL {}\ \BBA {} Hilgard%
}{%
Carter%
\ \protect \BOthers {.}}{%
{\protect \APACyear {2019}}%
}]{%
carter2019correcting}
\APACinsertmetastar {%
carter2019correcting}%
\begin{APACrefauthors}%
Carter, E\BPBI C.%
, Sch{\"o}nbrodt, F\BPBI D.%
, Gervais, W\BPBI M.%
\BCBL {}\ \BBA {} Hilgard, J.%
\end{APACrefauthors}%
\unskip\
\newblock
\APACrefYearMonthDay{2019}{}{}.
\newblock
{\BBOQ}\APACrefatitle {Correcting for bias in psychology: A comparison of
  meta-analytic methods} {Correcting for bias in psychology: A comparison of
  meta-analytic methods}.{\BBCQ}
\newblock
\APACjournalVolNumPages{Advances in Methods and Practices in Psychological
  Science}{2}{2}{115--144}.
\PrintBackRefs{\CurrentBib}

\bibitem [\protect \citeauthoryear {%
Castelnovo%
\ \protect \BOthers {.}}{%
Castelnovo%
\ \protect \BOthers {.}}{%
{\protect \APACyear {2022}}%
}]{%
Castelnovo.2022}
\APACinsertmetastar {%
Castelnovo.2022}%
\begin{APACrefauthors}%
Castelnovo, A.%
, Crupi, R.%
, Greco, G.%
, Regoli, D.%
, Penco, I\BPBI G.%
\BCBL {}\ \BBA {} Cosentini, A\BPBI C.%
\end{APACrefauthors}%
\unskip\
\newblock
\APACrefYearMonthDay{2022}{}{}.
\newblock
{\BBOQ}\APACrefatitle {A clarification of the nuances in the fairness metrics
  landscape} {A clarification of the nuances in the fairness metrics
  landscape}.{\BBCQ}
\newblock
\APACjournalVolNumPages{Scientific Reports}{12}{1}{1--21}.
\PrintBackRefs{\CurrentBib}

\bibitem [\protect \citeauthoryear {%
Caton%
\ \BBA {} Haas%
}{%
Caton%
\ \BBA {} Haas%
}{%
{\protect \APACyear {2020}}%
}]{%
https://doi.org/10.48550/arxiv.2010.04053}
\APACinsertmetastar {%
https://doi.org/10.48550/arxiv.2010.04053}%
\begin{APACrefauthors}%
Caton, S.%
\BCBT {}\ \BBA {} Haas, C.%
\end{APACrefauthors}%
\unskip\
\newblock
\APACrefYearMonthDay{2020}{}{}.
\newblock
{\BBOQ}\APACrefatitle {Fairness in Machine Learning: A Survey} {Fairness in
  machine learning: A survey}.{\BBCQ}
\newblock
\APACjournalVolNumPages{arXiv preprint arXiv:2010.04053}{}{}{}.
\PrintBackRefs{\CurrentBib}

\bibitem [\protect \citeauthoryear {%
Chambers%
}{%
Chambers%
}{%
{\protect \APACyear {2013}}%
}]{%
chambers2013registered}
\APACinsertmetastar {%
chambers2013registered}%
\begin{APACrefauthors}%
Chambers, C\BPBI D.%
\end{APACrefauthors}%
\unskip\
\newblock
\APACrefYearMonthDay{2013}{}{}.
\newblock
{\BBOQ}\APACrefatitle {Registered reports: a new publishing initiative at
  Cortex} {Registered reports: a new publishing initiative at cortex}.{\BBCQ}
\newblock
\APACjournalVolNumPages{Cortex}{49}{3}{609--610}.
\PrintBackRefs{\CurrentBib}

\bibitem [\protect \citeauthoryear {%
Cherian%
\ \BBA {} Cand{\`e}s%
}{%
Cherian%
\ \BBA {} Cand{\`e}s%
}{%
{\protect \APACyear {2023}}%
}]{%
cherian2023statistical}
\APACinsertmetastar {%
cherian2023statistical}%
\begin{APACrefauthors}%
Cherian, J\BPBI J.%
\BCBT {}\ \BBA {} Cand{\`e}s, E\BPBI J.%
\end{APACrefauthors}%
\unskip\
\newblock
\APACrefYearMonthDay{2023}{}{}.
\newblock
{\BBOQ}\APACrefatitle {Statistical Inference for Fairness Auditing}
  {Statistical inference for fairness auditing}.{\BBCQ}
\newblock
\APACjournalVolNumPages{arXiv preprint arXiv:2305.03712}{}{}{}.
\PrintBackRefs{\CurrentBib}

\bibitem [\protect \citeauthoryear {%
Cohen%
}{%
Cohen%
}{%
{\protect \APACyear {1969}}%
}]{%
cohen1969statistical}
\APACinsertmetastar {%
cohen1969statistical}%
\begin{APACrefauthors}%
Cohen, J.%
\end{APACrefauthors}%
\unskip\
\newblock
\APACrefYear{1969}.
\newblock
\APACrefbtitle {Statistical power analysis for the behavioral sciences}
  {Statistical power analysis for the behavioral sciences}.
\newblock
\APACaddressPublisher{}{Academic press}.
\PrintBackRefs{\CurrentBib}

\bibitem [\protect \citeauthoryear {%
Corbett-Davies%
, Pierson%
\BCBL {}\ \BBA {} Goel%
}{%
Corbett-Davies%
\ \protect \BOthers {.}}{%
{\protect \APACyear {2016}}%
}]{%
CorbettDavies.2016}
\APACinsertmetastar {%
CorbettDavies.2016}%
\begin{APACrefauthors}%
Corbett-Davies, S.%
, Pierson, E.%
\BCBL {}\ \BBA {} Goel, S.%
\end{APACrefauthors}%
\unskip\
\newblock
\APACrefYearMonthDay{2016}{}{}.
\newblock
\APACrefbtitle {A computer program used for bail and sentencing decisions was
  labeled biased against blacks. It's actually not that clear.} {A computer
  program used for bail and sentencing decisions was labeled biased against
  blacks. it's actually not that clear.}
\newblock
\begin{APACrefURL}
  [{16.09.2021}]\url{https://www.washingtonpost.com/news/monkey-cage/wp/2016/10/17/can-an-algorithm-be-racist-our-analysis-is-more-cautious-than-propublicas/}
  \end{APACrefURL}
\PrintBackRefs{\CurrentBib}

\bibitem [\protect \citeauthoryear {%
Cumming%
}{%
Cumming%
}{%
{\protect \APACyear {2014}}%
}]{%
cumming2014new}
\APACinsertmetastar {%
cumming2014new}%
\begin{APACrefauthors}%
Cumming, G.%
\end{APACrefauthors}%
\unskip\
\newblock
\APACrefYearMonthDay{2014}{}{}.
\newblock
{\BBOQ}\APACrefatitle {The new statistics: Why and how} {The new statistics:
  Why and how}.{\BBCQ}
\newblock
\APACjournalVolNumPages{Psychological science}{25}{1}{7--29}.
\PrintBackRefs{\CurrentBib}

\bibitem [\protect \citeauthoryear {%
Dimitrakakis%
, Liu%
, Parkes%
\BCBL {}\ \BBA {} Radanovic%
}{%
Dimitrakakis%
\ \protect \BOthers {.}}{%
{\protect \APACyear {2019}}%
}]{%
dimitrakakis2019bayesian}
\APACinsertmetastar {%
dimitrakakis2019bayesian}%
\begin{APACrefauthors}%
Dimitrakakis, C.%
, Liu, Y.%
, Parkes, D\BPBI C.%
\BCBL {}\ \BBA {} Radanovic, G.%
\end{APACrefauthors}%
\unskip\
\newblock
\APACrefYearMonthDay{2019}{}{}.
\newblock
{\BBOQ}\APACrefatitle {Bayesian fairness} {Bayesian fairness}.{\BBCQ}
\newblock
\BIn{} \APACrefbtitle {Proceedings of the AAAI Conference on Artificial
  Intelligence} {Proceedings of the aaai conference on artificial
  intelligence}\ (\BVOL~33, \BPGS\ 509--516).
\PrintBackRefs{\CurrentBib}

\bibitem [\protect \citeauthoryear {%
Ding%
, Hardt%
, Miller%
\BCBL {}\ \BBA {} Schmidt%
}{%
Ding%
\ \protect \BOthers {.}}{%
{\protect \APACyear {2021}}%
}]{%
ding2021retiring}
\APACinsertmetastar {%
ding2021retiring}%
\begin{APACrefauthors}%
Ding, F.%
, Hardt, M.%
, Miller, J.%
\BCBL {}\ \BBA {} Schmidt, L.%
\end{APACrefauthors}%
\unskip\
\newblock
\APACrefYearMonthDay{2021}{}{}.
\newblock
{\BBOQ}\APACrefatitle {Retiring adult: New datasets for fair machine learning}
  {Retiring adult: New datasets for fair machine learning}.{\BBCQ}
\newblock
\APACjournalVolNumPages{Advances in Neural Information Processing
  Systems}{34}{}{6478--6490}.
\PrintBackRefs{\CurrentBib}

\bibitem [\protect \citeauthoryear {%
Fanelli%
}{%
Fanelli%
}{%
{\protect \APACyear {2012}}%
}]{%
fanelli2012negative}
\APACinsertmetastar {%
fanelli2012negative}%
\begin{APACrefauthors}%
Fanelli, D.%
\end{APACrefauthors}%
\unskip\
\newblock
\APACrefYearMonthDay{2012}{}{}.
\newblock
{\BBOQ}\APACrefatitle {Negative results are disappearing from most disciplines
  and countries} {Negative results are disappearing from most disciplines and
  countries}.{\BBCQ}
\newblock
\APACjournalVolNumPages{Scientometrics}{90}{3}{891--904}.
\PrintBackRefs{\CurrentBib}

\bibitem [\protect \citeauthoryear {%
Fazelpour%
\ \BBA {} Lipton%
}{%
Fazelpour%
\ \BBA {} Lipton%
}{%
{\protect \APACyear {2020}}%
}]{%
Fazelpour.2020}
\APACinsertmetastar {%
Fazelpour.2020}%
\begin{APACrefauthors}%
Fazelpour, S.%
\BCBT {}\ \BBA {} Lipton, Z\BPBI C.%
\end{APACrefauthors}%
\unskip\
\newblock
\APACrefYearMonthDay{2020}{}{}.
\newblock
{\BBOQ}\APACrefatitle {Algorithmic Fairness from a Non-ideal Perspective}
  {Algorithmic fairness from a non-ideal perspective}.{\BBCQ}
\newblock
\BIn{} A.~Markham, J.~Powles, T.~Walsh\BCBL {}\ \BBA {} A\BPBI L.~Washington\
  (\BEDS), \APACrefbtitle {Proceedings of the AAAI/ACM Conference on AI,
  Ethics, and Society} {Proceedings of the aaai/acm conference on ai, ethics,
  and society}\ (\BPGS\ 57--63).
\newblock
\APACaddressPublisher{New York}{ACM}.
\PrintBackRefs{\CurrentBib}

\bibitem [\protect \citeauthoryear {%
Foulds%
, Islam%
, Keya%
\BCBL {}\ \BBA {} Pan%
}{%
Foulds%
\ \protect \BOthers {.}}{%
{\protect \APACyear {2020}}%
}]{%
foulds2020bayesian}
\APACinsertmetastar {%
foulds2020bayesian}%
\begin{APACrefauthors}%
Foulds, J\BPBI R.%
, Islam, R.%
, Keya, K\BPBI N.%
\BCBL {}\ \BBA {} Pan, S.%
\end{APACrefauthors}%
\unskip\
\newblock
\APACrefYearMonthDay{2020}{}{}.
\newblock
{\BBOQ}\APACrefatitle {Bayesian Modeling of Intersectional Fairness: The
  Variance of Bias} {Bayesian modeling of intersectional fairness: The variance
  of bias}.{\BBCQ}
\newblock
\BIn{} \APACrefbtitle {Proceedings of the 2020 SIAM International Conference on
  Data Mining} {Proceedings of the 2020 siam international conference on data
  mining}\ (\BPGS\ 424--432).
\PrintBackRefs{\CurrentBib}

\bibitem [\protect \citeauthoryear {%
Friedler%
, Scheidegger%
\BCBL {}\ \BBA {} Venkatasubramanian%
}{%
Friedler%
\ \protect \BOthers {.}}{%
{\protect \APACyear {2016}}%
}]{%
https://doi.org/10.48550/arxiv.1609.07236}
\APACinsertmetastar {%
https://doi.org/10.48550/arxiv.1609.07236}%
\begin{APACrefauthors}%
Friedler, S\BPBI A.%
, Scheidegger, C.%
\BCBL {}\ \BBA {} Venkatasubramanian, S.%
\end{APACrefauthors}%
\unskip\
\newblock
\APACrefYearMonthDay{2016}{}{}.
\newblock
{\BBOQ}\APACrefatitle {On the (im)possibility of fairness} {On the
  (im)possibility of fairness}.{\BBCQ}
\newblock
\APACjournalVolNumPages{arXiv preprint arXiv:1609.07236}{}{}{}.
\newblock
\begin{APACrefURL} \url{https://arxiv.org/abs/1609.07236} \end{APACrefURL}
\PrintBackRefs{\CurrentBib}

\bibitem [\protect \citeauthoryear {%
Friedler%
\ \protect \BOthers {.}}{%
Friedler%
\ \protect \BOthers {.}}{%
{\protect \APACyear {2018}}%
}]{%
friedler2018comparative}
\APACinsertmetastar {%
friedler2018comparative}%
\begin{APACrefauthors}%
Friedler, S\BPBI A.%
, Scheidegger, C.%
, Venkatasubramanian, S.%
, Choudhary, S.%
, Hamilton, E\BPBI P.%
\BCBL {}\ \BBA {} Roth, D.%
\end{APACrefauthors}%
\unskip\
\newblock
\APACrefYearMonthDay{2018}{}{}.
\newblock
\APACrefbtitle {A comparative study of fairness-enhancing interventions in
  machine learning.} {A comparative study of fairness-enhancing interventions
  in machine learning.}
\PrintBackRefs{\CurrentBib}

\bibitem [\protect \citeauthoryear {%
Gajane%
\ \BBA {} Pechenizkiy%
}{%
Gajane%
\ \BBA {} Pechenizkiy%
}{%
{\protect \APACyear {2017}}%
}]{%
https://doi.org/10.48550/arxiv.1710.03184}
\APACinsertmetastar {%
https://doi.org/10.48550/arxiv.1710.03184}%
\begin{APACrefauthors}%
Gajane, P.%
\BCBT {}\ \BBA {} Pechenizkiy, M.%
\end{APACrefauthors}%
\unskip\
\newblock
\APACrefYearMonthDay{2017}{}{}.
\newblock
{\BBOQ}\APACrefatitle {On Formalizing Fairness in Prediction with Machine
  Learning} {On formalizing fairness in prediction with machine
  learning}.{\BBCQ}
\newblock
\APACjournalVolNumPages{arXiv preprint arXiv:1710.03184}{}{}{}.
\PrintBackRefs{\CurrentBib}

\bibitem [\protect \citeauthoryear {%
Gebru%
\ \protect \BOthers {.}}{%
Gebru%
\ \protect \BOthers {.}}{%
{\protect \APACyear {2021}}%
}]{%
gebru2021datasheets}
\APACinsertmetastar {%
gebru2021datasheets}%
\begin{APACrefauthors}%
Gebru, T.%
, Morgenstern, J.%
, Vecchione, B.%
, Vaughan, J\BPBI W.%
, Wallach, H.%
, Iii, H\BPBI D.%
\BCBL {}\ \BBA {} Crawford, K.%
\end{APACrefauthors}%
\unskip\
\newblock
\APACrefYearMonthDay{2021}{}{}.
\newblock
{\BBOQ}\APACrefatitle {Datasheets for datasets} {Datasheets for
  datasets}.{\BBCQ}
\newblock
\APACjournalVolNumPages{Communications of the ACM}{64}{12}{86--92}.
\PrintBackRefs{\CurrentBib}

\bibitem [\protect \citeauthoryear {%
Green%
\ \BBA {} Viljoen%
}{%
Green%
\ \BBA {} Viljoen%
}{%
{\protect \APACyear {2020}}%
}]{%
Green.2020}
\APACinsertmetastar {%
Green.2020}%
\begin{APACrefauthors}%
Green, B.%
\BCBT {}\ \BBA {} Viljoen, S.%
\end{APACrefauthors}%
\unskip\
\newblock
\APACrefYearMonthDay{2020}{}{}.
\newblock
{\BBOQ}\APACrefatitle {Algorithmic realism} {Algorithmic realism}.{\BBCQ}
\newblock
\BIn{} M.~Hildebrandt, C.~Castillo, E.~Celis, S.~Ruggieri, L.~Taylor\BCBL {}\
  \BBA {} G.~Zanfir-Fortuna\ (\BEDS), \APACrefbtitle {Proceedings of the 2020
  Conference on Fairness, Accountability, and Transparency} {Proceedings of the
  2020 conference on fairness, accountability, and transparency}\ (\BPGS\
  19--31).
\newblock
\APACaddressPublisher{New York, NY, USA}{ACM}.
\PrintBackRefs{\CurrentBib}

\bibitem [\protect \citeauthoryear {%
Gundersen%
}{%
Gundersen%
}{%
{\protect \APACyear {2021}}%
}]{%
gundersen2021case}
\APACinsertmetastar {%
gundersen2021case}%
\begin{APACrefauthors}%
Gundersen, O\BPBI E.%
\end{APACrefauthors}%
\unskip\
\newblock
\APACrefYearMonthDay{2021}{}{}.
\newblock
{\BBOQ}\APACrefatitle {The case against registered reports} {The case against
  registered reports}.{\BBCQ}
\newblock
\APACjournalVolNumPages{AI Magazine}{42}{1}{88--92}.
\PrintBackRefs{\CurrentBib}

\bibitem [\protect \citeauthoryear {%
Hanna%
, Denton%
, Smart%
\BCBL {}\ \BBA {} Smith-Loud%
}{%
Hanna%
\ \protect \BOthers {.}}{%
{\protect \APACyear {2020}}%
}]{%
Hanna.2020}
\APACinsertmetastar {%
Hanna.2020}%
\begin{APACrefauthors}%
Hanna, A.%
, Denton, E.%
, Smart, A.%
\BCBL {}\ \BBA {} Smith-Loud, J.%
\end{APACrefauthors}%
\unskip\
\newblock
\APACrefYearMonthDay{2020}{}{}.
\newblock
{\BBOQ}\APACrefatitle {Towards a critical race methodology in algorithmic
  fairness} {Towards a critical race methodology in algorithmic
  fairness}.{\BBCQ}
\newblock
\BIn{} M.~Hildebrandt, C.~Castillo, E.~Celis, S.~Ruggieri, L.~Taylor\BCBL {}\
  \BBA {} G.~Zanfir-Fortuna\ (\BEDS), \APACrefbtitle {Proceedings of the 2020
  Conference on Fairness, Accountability, and Transparency} {Proceedings of the
  2020 conference on fairness, accountability, and transparency}\ (\BPGS\
  501--512).
\newblock
\APACaddressPublisher{New York}{ACM}.
\PrintBackRefs{\CurrentBib}

\bibitem [\protect \citeauthoryear {%
Head%
, Holman%
, Lanfear%
, Kahn%
\BCBL {}\ \BBA {} Jennions%
}{%
Head%
\ \protect \BOthers {.}}{%
{\protect \APACyear {2015}}%
}]{%
head2015extent}
\APACinsertmetastar {%
head2015extent}%
\begin{APACrefauthors}%
Head, M\BPBI L.%
, Holman, L.%
, Lanfear, R.%
, Kahn, A\BPBI T.%
\BCBL {}\ \BBA {} Jennions, M\BPBI D.%
\end{APACrefauthors}%
\unskip\
\newblock
\APACrefYearMonthDay{2015}{}{}.
\newblock
{\BBOQ}\APACrefatitle {The extent and consequences of p-hacking in science}
  {The extent and consequences of p-hacking in science}.{\BBCQ}
\newblock
\APACjournalVolNumPages{PLoS biology}{13}{3}{e1002106}.
\PrintBackRefs{\CurrentBib}

\bibitem [\protect \citeauthoryear {%
Heidari%
, Loi%
, Gummadi%
\BCBL {}\ \BBA {} Krause%
}{%
Heidari%
\ \protect \BOthers {.}}{%
{\protect \APACyear {2018}}%
}]{%
https://doi.org/10.48550/arxiv.1809.03400}
\APACinsertmetastar {%
https://doi.org/10.48550/arxiv.1809.03400}%
\begin{APACrefauthors}%
Heidari, H.%
, Loi, M.%
, Gummadi, K\BPBI P.%
\BCBL {}\ \BBA {} Krause, A.%
\end{APACrefauthors}%
\unskip\
\newblock
\APACrefYearMonthDay{2018}{}{}.
\newblock
{\BBOQ}\APACrefatitle {A Moral Framework for Understanding of Fair ML through
  Economic Models of Equality of Opportunity} {A moral framework for
  understanding of fair ml through economic models of equality of
  opportunity}.{\BBCQ}
\newblock
\APACjournalVolNumPages{arXiv preprint arXiv:1809.03400}{}{}{}.
\PrintBackRefs{\CurrentBib}

\bibitem [\protect \citeauthoryear {%
Hoffmann%
}{%
Hoffmann%
}{%
{\protect \APACyear {2019}}%
}]{%
Hoffmann.2019}
\APACinsertmetastar {%
Hoffmann.2019}%
\begin{APACrefauthors}%
Hoffmann, A\BPBI L.%
\end{APACrefauthors}%
\unskip\
\newblock
\APACrefYearMonthDay{2019}{}{}.
\newblock
{\BBOQ}\APACrefatitle {Where fairness fails: data, algorithms, and the limits
  of antidiscrimination discourse} {Where fairness fails: data, algorithms, and
  the limits of antidiscrimination discourse}.{\BBCQ}
\newblock
\APACjournalVolNumPages{Information, Communication {\&}
  Society}{22}{7}{900--915}.
\PrintBackRefs{\CurrentBib}

\bibitem [\protect \citeauthoryear {%
Holstein%
, Vaughan%
, Daum{\'e}%
, Dud{\'i}k%
\BCBL {}\ \BBA {} Wallach%
}{%
Holstein%
\ \protect \BOthers {.}}{%
{\protect \APACyear {2019}}%
}]{%
Holstein.2019}
\APACinsertmetastar {%
Holstein.2019}%
\begin{APACrefauthors}%
Holstein, K.%
, Vaughan, J\BPBI W.%
, Daum{\'e}, H., III%
, Dud{\'i}k, M.%
\BCBL {}\ \BBA {} Wallach, H.%
\end{APACrefauthors}%
\unskip\
\newblock
\APACrefYearMonthDay{2019}{}{}.
\newblock
{\BBOQ}\APACrefatitle {Improving fairness in machine learning systems: What do
  industry practitioners need?} {Improving fairness in machine learning
  systems: What do industry practitioners need?}{\BBCQ}
\newblock
\APACjournalVolNumPages{arXiv preprint arXiv:1812.05239}{}{}{}.
\PrintBackRefs{\CurrentBib}

\bibitem [\protect \citeauthoryear {%
Ioannidis%
}{%
Ioannidis%
}{%
{\protect \APACyear {2005}}%
}]{%
Ioannidis.2005}
\APACinsertmetastar {%
Ioannidis.2005}%
\begin{APACrefauthors}%
Ioannidis, J\BPBI P\BPBI A.%
\end{APACrefauthors}%
\unskip\
\newblock
\APACrefYearMonthDay{2005}{}{}.
\newblock
{\BBOQ}\APACrefatitle {Why most published research findings are false} {Why
  most published research findings are false}.{\BBCQ}
\newblock
\APACjournalVolNumPages{PLoS Medicine}{2}{8}{696--701}.
\PrintBackRefs{\CurrentBib}

\bibitem [\protect \citeauthoryear {%
Ji%
, Smyth%
\BCBL {}\ \BBA {} Steyvers%
}{%
Ji%
\ \protect \BOthers {.}}{%
{\protect \APACyear {2020}}%
}]{%
ji2020can}
\APACinsertmetastar {%
ji2020can}%
\begin{APACrefauthors}%
Ji, D.%
, Smyth, P.%
\BCBL {}\ \BBA {} Steyvers, M.%
\end{APACrefauthors}%
\unskip\
\newblock
\APACrefYearMonthDay{2020}{}{}.
\newblock
{\BBOQ}\APACrefatitle {Can i trust my fairness metric? assessing fairness with
  unlabeled data and bayesian inference} {Can i trust my fairness metric?
  assessing fairness with unlabeled data and bayesian inference}.{\BBCQ}
\newblock
\APACjournalVolNumPages{Advances in Neural Information Processing
  Systems}{33}{}{18600--18612}.
\PrintBackRefs{\CurrentBib}

\bibitem [\protect \citeauthoryear {%
Jo%
, Sohn%
\BCBL {}\ \BBA {} Lee%
}{%
Jo%
\ \protect \BOthers {.}}{%
{\protect \APACyear {2022}}%
}]{%
Jo.2022}
\APACinsertmetastar {%
Jo.2022}%
\begin{APACrefauthors}%
Jo, C.%
, Sohn, J\BHBI Y.%
\BCBL {}\ \BBA {} Lee, K.%
\end{APACrefauthors}%
\unskip\
\newblock
\APACrefYearMonthDay{2022}{}{}.
\newblock
{\BBOQ}\APACrefatitle {Breaking Fair Binary Classification with Optimal
  Flipping Attacks} {Breaking fair binary classification with optimal flipping
  attacks}.{\BBCQ}
\newblock
\APACjournalVolNumPages{arXiv preprint arXiv:2204.05472}{}{}{}.
\PrintBackRefs{\CurrentBib}

\bibitem [\protect \citeauthoryear {%
Jobin%
, Ienca%
\BCBL {}\ \BBA {} Vayena%
}{%
Jobin%
\ \protect \BOthers {.}}{%
{\protect \APACyear {2019}}%
}]{%
Jobin.2019}
\APACinsertmetastar {%
Jobin.2019}%
\begin{APACrefauthors}%
Jobin, A.%
, Ienca, M.%
\BCBL {}\ \BBA {} Vayena, E.%
\end{APACrefauthors}%
\unskip\
\newblock
\APACrefYearMonthDay{2019}{}{}.
\newblock
{\BBOQ}\APACrefatitle {The global landscape of AI ethics guidelines} {The
  global landscape of ai ethics guidelines}.{\BBCQ}
\newblock
\APACjournalVolNumPages{Nature Machine Intelligence}{1}{9}{389--399}.
\PrintBackRefs{\CurrentBib}

\bibitem [\protect \citeauthoryear {%
John%
, Loewenstein%
\BCBL {}\ \BBA {} Prelec%
}{%
John%
\ \protect \BOthers {.}}{%
{\protect \APACyear {2012}}%
}]{%
john2012measuring}
\APACinsertmetastar {%
john2012measuring}%
\begin{APACrefauthors}%
John, L\BPBI K.%
, Loewenstein, G.%
\BCBL {}\ \BBA {} Prelec, D.%
\end{APACrefauthors}%
\unskip\
\newblock
\APACrefYearMonthDay{2012}{}{}.
\newblock
{\BBOQ}\APACrefatitle {Measuring the prevalence of questionable research
  practices with incentives for truth telling} {Measuring the prevalence of
  questionable research practices with incentives for truth telling}.{\BBCQ}
\newblock
\APACjournalVolNumPages{Psychological science}{23}{5}{524--532}.
\PrintBackRefs{\CurrentBib}

\bibitem [\protect \citeauthoryear {%
John-Mathews%
\ \BBA {} Cardon%
}{%
John-Mathews%
\ \BBA {} Cardon%
}{%
{\protect \APACyear {2022}}%
}]{%
john2022crisis}
\APACinsertmetastar {%
john2022crisis}%
\begin{APACrefauthors}%
John-Mathews, J\BHBI M.%
\BCBT {}\ \BBA {} Cardon, D.%
\end{APACrefauthors}%
\unskip\
\newblock
\APACrefYearMonthDay{2022}{}{}.
\newblock
{\BBOQ}\APACrefatitle {The Crisis of Social Categories in the Age of AI} {The
  crisis of social categories in the age of ai}.{\BBCQ}
\newblock
\APACjournalVolNumPages{Sociologica}{16}{3}{5--16}.
\PrintBackRefs{\CurrentBib}

\bibitem [\protect \citeauthoryear {%
John-Mathews%
, Mourat%
, Ricci%
\BCBL {}\ \BBA {} Cr{\'{e}}pel%
}{%
John-Mathews%
\ \protect \BOthers {.}}{%
{\protect \APACyear {2023}}%
}]{%
JohnMathews2023}
\APACinsertmetastar {%
JohnMathews2023}%
\begin{APACrefauthors}%
John-Mathews, J\BHBI M.%
, Mourat, R\BPBI D.%
, Ricci, D.%
\BCBL {}\ \BBA {} Cr{\'{e}}pel, M.%
\end{APACrefauthors}%
\unskip\
\newblock
\APACrefYearMonthDay{2023}{{\APACmonth{07}}}{}.
\newblock
{\BBOQ}\APACrefatitle {Re-enacting machine learning practices to enquire into
  the moral issues they pose} {Re-enacting machine learning practices to
  enquire into the moral issues they pose}.{\BBCQ}
\newblock
\APACjournalVolNumPages{Convergence: The International Journal of Research into
  New Media Technologies}{}{}{}.
\newblock
\begin{APACrefURL} \url{https://doi.org/10.1177/13548565231174584}
  \end{APACrefURL}
\newblock
\begin{APACrefDOI} \doi{10.1177/13548565231174584} \end{APACrefDOI}
\PrintBackRefs{\CurrentBib}

\bibitem [\protect \citeauthoryear {%
Kearns%
, Neel%
, Roth%
\BCBL {}\ \BBA {} Wu%
}{%
Kearns%
\ \protect \BOthers {.}}{%
{\protect \APACyear {2018}}%
}]{%
pmlr-v80-kearns18a}
\APACinsertmetastar {%
pmlr-v80-kearns18a}%
\begin{APACrefauthors}%
Kearns, M.%
, Neel, S.%
, Roth, A.%
\BCBL {}\ \BBA {} Wu, Z\BPBI S.%
\end{APACrefauthors}%
\unskip\
\newblock
\APACrefYearMonthDay{2018}{10--15 Jul}{}.
\newblock
{\BBOQ}\APACrefatitle {Preventing Fairness Gerrymandering: Auditing and
  Learning for Subgroup Fairness} {Preventing fairness gerrymandering: Auditing
  and learning for subgroup fairness}.{\BBCQ}
\newblock
\BIn{} J.~Dy\ \BBA {} A.~Krause\ (\BEDS), \APACrefbtitle {Proceedings of the
  35th International Conference on Machine Learning} {Proceedings of the 35th
  international conference on machine learning}\ (\BVOL~80, \BPGS\ 2564--2572).
\newblock
\APACaddressPublisher{}{PMLR}.
\PrintBackRefs{\CurrentBib}

\bibitem [\protect \citeauthoryear {%
Kerr%
}{%
Kerr%
}{%
{\protect \APACyear {1998}}%
}]{%
kerr1998harking}
\APACinsertmetastar {%
kerr1998harking}%
\begin{APACrefauthors}%
Kerr, N\BPBI L.%
\end{APACrefauthors}%
\unskip\
\newblock
\APACrefYearMonthDay{1998}{}{}.
\newblock
{\BBOQ}\APACrefatitle {HARKing: Hypothesizing after the results are known}
  {Harking: Hypothesizing after the results are known}.{\BBCQ}
\newblock
\APACjournalVolNumPages{Personality and social psychology
  review}{2}{3}{196--217}.
\PrintBackRefs{\CurrentBib}

\bibitem [\protect \citeauthoryear {%
Kleinberg%
, Mullainathan%
\BCBL {}\ \BBA {} Raghavan%
}{%
Kleinberg%
\ \protect \BOthers {.}}{%
{\protect \APACyear {2016}}%
}]{%
kleinberg2016inherent}
\APACinsertmetastar {%
kleinberg2016inherent}%
\begin{APACrefauthors}%
Kleinberg, J.%
, Mullainathan, S.%
\BCBL {}\ \BBA {} Raghavan, M.%
\end{APACrefauthors}%
\unskip\
\newblock
\APACrefYearMonthDay{2016}{}{}.
\newblock
{\BBOQ}\APACrefatitle {Inherent trade-offs in the fair determination of risk
  scores} {Inherent trade-offs in the fair determination of risk
  scores}.{\BBCQ}
\newblock
\APACjournalVolNumPages{arXiv preprint arXiv:1609.05807}{}{}{}.
\PrintBackRefs{\CurrentBib}

\bibitem [\protect \citeauthoryear {%
Lum%
, Zhang%
\BCBL {}\ \BBA {} Bower%
}{%
Lum%
\ \protect \BOthers {.}}{%
{\protect \APACyear {2022}}%
}]{%
Lum_2022}
\APACinsertmetastar {%
Lum_2022}%
\begin{APACrefauthors}%
Lum, K.%
, Zhang, Y.%
\BCBL {}\ \BBA {} Bower, A.%
\end{APACrefauthors}%
\unskip\
\newblock
\APACrefYearMonthDay{2022}{jun}{}.
\newblock
{\BBOQ}\APACrefatitle {De-biasing {\textquotedblleft}bias{\textquotedblright}
  measurement} {De-biasing {\textquotedblleft}bias{\textquotedblright}
  measurement}.{\BBCQ}
\newblock
\BIn{} \APACrefbtitle {2022 {ACM} Conference on Fairness, Accountability, and
  Transparency.} {2022 {ACM} conference on fairness, accountability, and
  transparency.}
\newblock
\APACaddressPublisher{}{{ACM}}.
\newblock
\begin{APACrefURL} \url{https://doi.org/10.1145%2F3531146.3533105}
  \end{APACrefURL}
\newblock
\begin{APACrefDOI} \doi{10.1145/3531146.3533105} \end{APACrefDOI}
\PrintBackRefs{\CurrentBib}

\bibitem [\protect \citeauthoryear {%
Lumley%
, Diehr%
, Emerson%
\BCBL {}\ \BBA {} Chen%
}{%
Lumley%
\ \protect \BOthers {.}}{%
{\protect \APACyear {2002}}%
}]{%
lumley2002importance}
\APACinsertmetastar {%
lumley2002importance}%
\begin{APACrefauthors}%
Lumley, T.%
, Diehr, P.%
, Emerson, S.%
\BCBL {}\ \BBA {} Chen, L.%
\end{APACrefauthors}%
\unskip\
\newblock
\APACrefYearMonthDay{2002}{}{}.
\newblock
{\BBOQ}\APACrefatitle {The importance of the normality assumption in large
  public health data sets} {The importance of the normality assumption in large
  public health data sets}.{\BBCQ}
\newblock
\APACjournalVolNumPages{Annual review of public health}{23}{1}{151--169}.
\PrintBackRefs{\CurrentBib}

\bibitem [\protect \citeauthoryear {%
Madaio%
, Stark%
, {Wortman Vaughan}%
\BCBL {}\ \BBA {} Wallach%
}{%
Madaio%
, Stark%
, {Wortman Vaughan}%
\BCBL {}\ \BBA {} Wallach%
}{%
{\protect \APACyear {2020}}%
}]{%
Madaio.2020}
\APACinsertmetastar {%
Madaio.2020}%
\begin{APACrefauthors}%
Madaio, M\BPBI A.%
, Stark, L.%
, {Wortman Vaughan}, J.%
\BCBL {}\ \BBA {} Wallach, H.%
\end{APACrefauthors}%
\unskip\
\newblock
\APACrefYearMonthDay{2020}{}{}.
\newblock
{\BBOQ}\APACrefatitle {Co-Designing Checklists to Understand Organizational
  Challenges and Opportunities around Fairness in AI} {Co-designing checklists
  to understand organizational challenges and opportunities around fairness in
  ai}.{\BBCQ}
\newblock
\BIn{} R.~Bernhaupt\ \BOthers {.}\ (\BEDS), \APACrefbtitle {Proceedings of the
  2020 CHI Conference on Human Factors in Computing Systems} {Proceedings of
  the 2020 chi conference on human factors in computing systems}\ (\BPGS\
  1--14).
\newblock
\APACaddressPublisher{New York, NY, USA}{ACM}.
\PrintBackRefs{\CurrentBib}

\bibitem [\protect \citeauthoryear {%
Madaio%
, Stark%
, Wortman~Vaughan%
\BCBL {}\ \BBA {} Wallach%
}{%
Madaio%
, Stark%
, Wortman~Vaughan%
\BCBL {}\ \BBA {} Wallach%
}{%
{\protect \APACyear {2020}}%
}]{%
madaio2020co}
\APACinsertmetastar {%
madaio2020co}%
\begin{APACrefauthors}%
Madaio, M\BPBI A.%
, Stark, L.%
, Wortman~Vaughan, J.%
\BCBL {}\ \BBA {} Wallach, H.%
\end{APACrefauthors}%
\unskip\
\newblock
\APACrefYearMonthDay{2020}{}{}.
\newblock
{\BBOQ}\APACrefatitle {Co-designing checklists to understand organizational
  challenges and opportunities around fairness in AI} {Co-designing checklists
  to understand organizational challenges and opportunities around fairness in
  ai}.{\BBCQ}
\newblock
\BIn{} \APACrefbtitle {Proceedings of the 2020 CHI conference on human factors
  in computing systems} {Proceedings of the 2020 chi conference on human
  factors in computing systems}\ (\BPGS\ 1--14).
\PrintBackRefs{\CurrentBib}

\bibitem [\protect \citeauthoryear {%
Mehrabi%
, Morstatter%
, Saxena%
, Lerman%
\BCBL {}\ \BBA {} Galstyan%
}{%
Mehrabi%
\ \protect \BOthers {.}}{%
{\protect \APACyear {2019}}%
}]{%
Mehrabi.2019}
\APACinsertmetastar {%
Mehrabi.2019}%
\begin{APACrefauthors}%
Mehrabi, N.%
, Morstatter, F.%
, Saxena, N.%
, Lerman, K.%
\BCBL {}\ \BBA {} Galstyan, A.%
\end{APACrefauthors}%
\unskip\
\newblock
\APACrefYearMonthDay{2019}{}{}.
\newblock
{\BBOQ}\APACrefatitle {A Survey on Bias and Fairness in Machine Learning} {A
  survey on bias and fairness in machine learning}.{\BBCQ}
\newblock
\APACjournalVolNumPages{arXiv preprint arXiv:1908.09635}{}{}{}.
\PrintBackRefs{\CurrentBib}

\bibitem [\protect \citeauthoryear {%
Mehta%
}{%
Mehta%
}{%
{\protect \APACyear {2019}}%
}]{%
Mehta.2019}
\APACinsertmetastar {%
Mehta.2019}%
\begin{APACrefauthors}%
Mehta, D.%
\end{APACrefauthors}%
\unskip\
\newblock
\APACrefYearMonthDay{2019}{}{}.
\newblock
\APACrefbtitle {Highlight negative results to improve science.} {Highlight
  negative results to improve science.}
\newblock
\begin{APACrefURL}
  [{21.10.2022}]\url{https://www.nature.com/articles/d41586-019-02960-3}
  \end{APACrefURL}
\PrintBackRefs{\CurrentBib}

\bibitem [\protect \citeauthoryear {%
M.~Mitchell%
\ \protect \BOthers {.}}{%
M.~Mitchell%
\ \protect \BOthers {.}}{%
{\protect \APACyear {2019}}%
}]{%
mitchell2019model}
\APACinsertmetastar {%
mitchell2019model}%
\begin{APACrefauthors}%
Mitchell, M.%
, Wu, S.%
, Zaldivar, A.%
, Barnes, P.%
, Vasserman, L.%
, Hutchinson, B.%
\BDBL {}Gebru, T.%
\end{APACrefauthors}%
\unskip\
\newblock
\APACrefYearMonthDay{2019}{}{}.
\newblock
{\BBOQ}\APACrefatitle {Model cards for model reporting} {Model cards for model
  reporting}.{\BBCQ}
\newblock
\BIn{} \APACrefbtitle {Proceedings of the conference on fairness,
  accountability, and transparency} {Proceedings of the conference on fairness,
  accountability, and transparency}\ (\BPGS\ 220--229).
\PrintBackRefs{\CurrentBib}

\bibitem [\protect \citeauthoryear {%
S.~Mitchell%
, Potash%
, Barocas%
, D'Amour%
\BCBL {}\ \BBA {} Lum%
}{%
S.~Mitchell%
\ \protect \BOthers {.}}{%
{\protect \APACyear {2021}}%
}]{%
Mitchell.2021}
\APACinsertmetastar {%
Mitchell.2021}%
\begin{APACrefauthors}%
Mitchell, S.%
, Potash, E.%
, Barocas, S.%
, D'Amour, A.%
\BCBL {}\ \BBA {} Lum, K.%
\end{APACrefauthors}%
\unskip\
\newblock
\APACrefYearMonthDay{2021}{}{}.
\newblock
{\BBOQ}\APACrefatitle {Algorithmic Fairness: Choices, Assumptions, and
  Definitions} {Algorithmic fairness: Choices, assumptions, and
  definitions}.{\BBCQ}
\newblock
\APACjournalVolNumPages{Annual Review of Statistics and Its
  Application}{8}{1}{141--163}.
\PrintBackRefs{\CurrentBib}

\bibitem [\protect \citeauthoryear {%
Newcombe%
}{%
Newcombe%
}{%
{\protect \APACyear {1998}}%
}]{%
newcombe1998two}
\APACinsertmetastar {%
newcombe1998two}%
\begin{APACrefauthors}%
Newcombe, R\BPBI G.%
\end{APACrefauthors}%
\unskip\
\newblock
\APACrefYearMonthDay{1998}{}{}.
\newblock
{\BBOQ}\APACrefatitle {Two-sided confidence intervals for the single
  proportion: comparison of seven methods} {Two-sided confidence intervals for
  the single proportion: comparison of seven methods}.{\BBCQ}
\newblock
\APACjournalVolNumPages{Statistics in medicine}{17}{8}{857--872}.
\PrintBackRefs{\CurrentBib}

\bibitem [\protect \citeauthoryear {%
Nosek%
, Spies%
\BCBL {}\ \BBA {} Motyl%
}{%
Nosek%
\ \protect \BOthers {.}}{%
{\protect \APACyear {2012}}%
}]{%
nosek2012scientific}
\APACinsertmetastar {%
nosek2012scientific}%
\begin{APACrefauthors}%
Nosek, B\BPBI A.%
, Spies, J\BPBI R.%
\BCBL {}\ \BBA {} Motyl, M.%
\end{APACrefauthors}%
\unskip\
\newblock
\APACrefYearMonthDay{2012}{}{}.
\newblock
{\BBOQ}\APACrefatitle {Scientific utopia: II. Restructuring incentives and
  practices to promote truth over publishability} {Scientific utopia: Ii.
  restructuring incentives and practices to promote truth over
  publishability}.{\BBCQ}
\newblock
\APACjournalVolNumPages{Perspectives on Psychological Science}{7}{6}{615--631}.
\PrintBackRefs{\CurrentBib}

\bibitem [\protect \citeauthoryear {%
{Open Science Collaboration}%
}{%
{Open Science Collaboration}%
}{%
{\protect \APACyear {2015}}%
}]{%
open2015estimating}
\APACinsertmetastar {%
open2015estimating}%
\begin{APACrefauthors}%
{Open Science Collaboration}.%
\end{APACrefauthors}%
\unskip\
\newblock
\APACrefYearMonthDay{2015}{}{}.
\newblock
{\BBOQ}\APACrefatitle {Estimating the reproducibility of psychological science}
  {Estimating the reproducibility of psychological science}.{\BBCQ}
\newblock
\APACjournalVolNumPages{Science}{349}{6251}{aac4716}.
\PrintBackRefs{\CurrentBib}

\bibitem [\protect \citeauthoryear {%
P~Simmons%
, D~Nelson%
\BCBL {}\ \BBA {} Simonsohn%
}{%
P~Simmons%
\ \protect \BOthers {.}}{%
{\protect \APACyear {2021}}%
}]{%
p2021pre}
\APACinsertmetastar {%
p2021pre}%
\begin{APACrefauthors}%
P~Simmons, J.%
, D~Nelson, L.%
\BCBL {}\ \BBA {} Simonsohn, U.%
\end{APACrefauthors}%
\unskip\
\newblock
\APACrefYearMonthDay{2021}{}{}.
\newblock
{\BBOQ}\APACrefatitle {Pre-registration: Why and how} {Pre-registration: Why
  and how}.{\BBCQ}
\newblock
\APACjournalVolNumPages{Journal of Consumer Psychology}{31}{1}{151--162}.
\PrintBackRefs{\CurrentBib}

\bibitem [\protect \citeauthoryear {%
Roy%
\ \BBA {} Mohapatra%
}{%
Roy%
\ \BBA {} Mohapatra%
}{%
{\protect \APACyear {2023}}%
}]{%
roy2023fairness}
\APACinsertmetastar {%
roy2023fairness}%
\begin{APACrefauthors}%
Roy, A.%
\BCBT {}\ \BBA {} Mohapatra, P.%
\end{APACrefauthors}%
\unskip\
\newblock
\APACrefYearMonthDay{2023}{}{}.
\newblock
{\BBOQ}\APACrefatitle {Fairness Uncertainty Quantification: How certain are you
  that the model is fair?} {Fairness uncertainty quantification: How certain
  are you that the model is fair?}{\BBCQ}
\newblock
\APACjournalVolNumPages{arXiv preprint arXiv:2304.13950}{}{}{}.
\PrintBackRefs{\CurrentBib}

\bibitem [\protect \citeauthoryear {%
Ruf%
\ \BBA {} Detyniecki%
}{%
Ruf%
\ \BBA {} Detyniecki%
}{%
{\protect \APACyear {2021}}%
}]{%
https://doi.org/10.48550/arxiv.2102.08453}
\APACinsertmetastar {%
https://doi.org/10.48550/arxiv.2102.08453}%
\begin{APACrefauthors}%
Ruf, B.%
\BCBT {}\ \BBA {} Detyniecki, M.%
\end{APACrefauthors}%
\unskip\
\newblock
\APACrefYearMonthDay{2021}{}{}.
\newblock
{\BBOQ}\APACrefatitle {Towards the Right Kind of Fairness in AI} {Towards the
  right kind of fairness in ai}.{\BBCQ}
\newblock
\APACjournalVolNumPages{arXiv preprint arXiv:2102.08453}{}{}{}.
\PrintBackRefs{\CurrentBib}

\bibitem [\protect \citeauthoryear {%
Saravanakumar%
}{%
Saravanakumar%
}{%
{\protect \APACyear {2021}}%
}]{%
Saravanakumar.2021}
\APACinsertmetastar {%
Saravanakumar.2021}%
\begin{APACrefauthors}%
Saravanakumar, K\BPBI K.%
\end{APACrefauthors}%
\unskip\
\newblock
\APACrefYearMonthDay{2021}{}{}.
\newblock
{\BBOQ}\APACrefatitle {The Impossibility Theorem of Machine Fairness -- A
  Causal Perspective} {The impossibility theorem of machine fairness -- a
  causal perspective}.{\BBCQ}
\newblock
\APACjournalVolNumPages{arXiv preprint arXiv:2007.06024}{}{}{}.
\PrintBackRefs{\CurrentBib}

\bibitem [\protect \citeauthoryear {%
Selbst%
, boyd%
, Friedler%
, Venkatasubramanian%
\BCBL {}\ \BBA {} Vertesi%
}{%
Selbst%
\ \protect \BOthers {.}}{%
{\protect \APACyear {2018}}%
}]{%
Selbst.2018b}
\APACinsertmetastar {%
Selbst.2018b}%
\begin{APACrefauthors}%
Selbst, A\BPBI D.%
, boyd, d.%
, Friedler, S\BPBI A.%
, Venkatasubramanian, S.%
\BCBL {}\ \BBA {} Vertesi, J.%
\end{APACrefauthors}%
\unskip\
\newblock
\APACrefYearMonthDay{2018}{}{}.
\newblock
{\BBOQ}\APACrefatitle {Fairness and Abstraction in Sociotechnical Systems}
  {Fairness and abstraction in sociotechnical systems}.{\BBCQ}
\newblock
\APACjournalVolNumPages{ACT Conference on Fairness, Accountability, and
  Transparency (FAT)}{1}{1}{1--17}.
\PrintBackRefs{\CurrentBib}

\bibitem [\protect \citeauthoryear {%
Shaffer%
}{%
Shaffer%
}{%
{\protect \APACyear {1995}}%
}]{%
shaffer1995multiple}
\APACinsertmetastar {%
shaffer1995multiple}%
\begin{APACrefauthors}%
Shaffer, J\BPBI P.%
\end{APACrefauthors}%
\unskip\
\newblock
\APACrefYearMonthDay{1995}{}{}.
\newblock
{\BBOQ}\APACrefatitle {Multiple hypothesis testing} {Multiple hypothesis
  testing}.{\BBCQ}
\newblock
\APACjournalVolNumPages{Annual review of psychology}{46}{1}{561--584}.
\PrintBackRefs{\CurrentBib}

\bibitem [\protect \citeauthoryear {%
Sharma%
\ \protect \BOthers {.}}{%
Sharma%
\ \protect \BOthers {.}}{%
{\protect \APACyear {2022}}%
}]{%
Sharma.2022}
\APACinsertmetastar {%
Sharma.2022}%
\begin{APACrefauthors}%
Sharma, R.%
, Kaushik, M.%
, Peious, S\BPBI A.%
, Bertl, M.%
, Vidyarthi, A.%
, Kumar, A.%
\BCBL {}\ \BBA {} Draheim, D.%
\end{APACrefauthors}%
\unskip\
\newblock
\APACrefYearMonthDay{2022}{}{}.
\newblock
{\BBOQ}\APACrefatitle {Detecting Simpson's Paradox: A Step Towards Fairness in
  Machine Learning} {Detecting simpson's paradox: A step towards fairness in
  machine learning}.{\BBCQ}
\newblock
\BIn{} S.~Chiusano\ \BOthers {.}\ (\BEDS), \APACrefbtitle {New Trends in
  Database and Information Systems} {New trends in database and information
  systems}\ (\BPGS\ 67--76).
\newblock
\APACaddressPublisher{Cham}{{Springer International Publishing}}.
\PrintBackRefs{\CurrentBib}

\bibitem [\protect \citeauthoryear {%
Solans%
, Biggio%
\BCBL {}\ \BBA {} Castillo%
}{%
Solans%
\ \protect \BOthers {.}}{%
{\protect \APACyear {2020}}%
}]{%
https://doi.org/10.48550/arxiv.2004.07401}
\APACinsertmetastar {%
https://doi.org/10.48550/arxiv.2004.07401}%
\begin{APACrefauthors}%
Solans, D.%
, Biggio, B.%
\BCBL {}\ \BBA {} Castillo, C.%
\end{APACrefauthors}%
\unskip\
\newblock
\APACrefYearMonthDay{2020}{}{}.
\newblock
{\BBOQ}\APACrefatitle {Poisoning Attacks on Algorithmic Fairness} {Poisoning
  attacks on algorithmic fairness}.{\BBCQ}
\newblock
\APACjournalVolNumPages{arXiv preprint arXiv:2004.07401}{}{}{}.
\PrintBackRefs{\CurrentBib}

\bibitem [\protect \citeauthoryear {%
Speicher%
\ \protect \BOthers {.}}{%
Speicher%
\ \protect \BOthers {.}}{%
{\protect \APACyear {2018}}%
}]{%
speicher2018unified}
\APACinsertmetastar {%
speicher2018unified}%
\begin{APACrefauthors}%
Speicher, T.%
, Heidari, H.%
, Grgic-Hlaca, N.%
, Gummadi, K\BPBI P.%
, Singla, A.%
, Weller, A.%
\BCBL {}\ \BBA {} Zafar, M\BPBI B.%
\end{APACrefauthors}%
\unskip\
\newblock
\APACrefYearMonthDay{2018}{}{}.
\newblock
{\BBOQ}\APACrefatitle {A unified approach to quantifying algorithmic
  unfairness: Measuring individual \&group unfairness via inequality indices}
  {A unified approach to quantifying algorithmic unfairness: Measuring
  individual \&group unfairness via inequality indices}.{\BBCQ}
\newblock
\BIn{} \APACrefbtitle {Proceedings of the 24th ACM SIGKDD international
  conference on knowledge discovery \& data mining} {Proceedings of the 24th
  acm sigkdd international conference on knowledge discovery \& data mining}\
  (\BPGS\ 2239--2248).
\PrintBackRefs{\CurrentBib}

\bibitem [\protect \citeauthoryear {%
Stefan%
\ \BBA {} Sch{\"o}nbrodt%
}{%
Stefan%
\ \BBA {} Sch{\"o}nbrodt%
}{%
{\protect \APACyear {2023}}%
}]{%
stefan2023big}
\APACinsertmetastar {%
stefan2023big}%
\begin{APACrefauthors}%
Stefan, A\BPBI M.%
\BCBT {}\ \BBA {} Sch{\"o}nbrodt, F\BPBI D.%
\end{APACrefauthors}%
\unskip\
\newblock
\APACrefYearMonthDay{2023}{}{}.
\newblock
{\BBOQ}\APACrefatitle {Big little lies: A compendium and simulation of
  p-hacking strategies} {Big little lies: A compendium and simulation of
  p-hacking strategies}.{\BBCQ}
\newblock
\APACjournalVolNumPages{Royal Society Open Science}{10}{2}{220346}.
\PrintBackRefs{\CurrentBib}

\bibitem [\protect \citeauthoryear {%
Sullivan%
\ \BBA {} Feinn%
}{%
Sullivan%
\ \BBA {} Feinn%
}{%
{\protect \APACyear {2012}}%
}]{%
sullivan2012using}
\APACinsertmetastar {%
sullivan2012using}%
\begin{APACrefauthors}%
Sullivan, G\BPBI M.%
\BCBT {}\ \BBA {} Feinn, R.%
\end{APACrefauthors}%
\unskip\
\newblock
\APACrefYearMonthDay{2012}{}{}.
\newblock
{\BBOQ}\APACrefatitle {Using effect size—or why the P value is not enough}
  {Using effect size—or why the p value is not enough}.{\BBCQ}
\newblock
\APACjournalVolNumPages{Journal of graduate medical education}{4}{3}{279--282}.
\PrintBackRefs{\CurrentBib}

\bibitem [\protect \citeauthoryear {%
Tukey%
\ \protect \BOthers {.}}{%
Tukey%
\ \protect \BOthers {.}}{%
{\protect \APACyear {1977}}%
}]{%
tukey1977exploratory}
\APACinsertmetastar {%
tukey1977exploratory}%
\begin{APACrefauthors}%
Tukey, J\BPBI W.%
\BCBT {}\ \BOthersPeriod {.}
\end{APACrefauthors}%
\unskip\
\newblock
\APACrefYear{1977}.
\newblock
\APACrefbtitle {Exploratory data analysis} {Exploratory data analysis}\
  (\BVOL~2).
\newblock
\APACaddressPublisher{}{Reading, MA}.
\PrintBackRefs{\CurrentBib}

\bibitem [\protect \citeauthoryear {%
Van't~Veer%
\ \BBA {} Giner-Sorolla%
}{%
Van't~Veer%
\ \BBA {} Giner-Sorolla%
}{%
{\protect \APACyear {2016}}%
}]{%
van2016pre}
\APACinsertmetastar {%
van2016pre}%
\begin{APACrefauthors}%
Van't~Veer, A\BPBI E.%
\BCBT {}\ \BBA {} Giner-Sorolla, R.%
\end{APACrefauthors}%
\unskip\
\newblock
\APACrefYearMonthDay{2016}{}{}.
\newblock
{\BBOQ}\APACrefatitle {Pre-registration in social psychology—A discussion and
  suggested template} {Pre-registration in social psychology—a discussion and
  suggested template}.{\BBCQ}
\newblock
\APACjournalVolNumPages{Journal of experimental social
  psychology}{67}{}{2--12}.
\PrintBackRefs{\CurrentBib}

\bibitem [\protect \citeauthoryear {%
Wagenmakers%
, Wetzels%
, Borsboom%
, van~der Maas%
\BCBL {}\ \BBA {} Kievit%
}{%
Wagenmakers%
\ \protect \BOthers {.}}{%
{\protect \APACyear {2012}}%
}]{%
wagenmakers2012agenda}
\APACinsertmetastar {%
wagenmakers2012agenda}%
\begin{APACrefauthors}%
Wagenmakers, E\BHBI J.%
, Wetzels, R.%
, Borsboom, D.%
, van~der Maas, H\BPBI L.%
\BCBL {}\ \BBA {} Kievit, R\BPBI A.%
\end{APACrefauthors}%
\unskip\
\newblock
\APACrefYearMonthDay{2012}{}{}.
\newblock
{\BBOQ}\APACrefatitle {An agenda for purely confirmatory research} {An agenda
  for purely confirmatory research}.{\BBCQ}
\newblock
\APACjournalVolNumPages{Perspectives on psychological science}{7}{6}{632--638}.
\PrintBackRefs{\CurrentBib}

\bibitem [\protect \citeauthoryear {%
Wainer%
}{%
Wainer%
}{%
{\protect \APACyear {2007}}%
}]{%
wainer2007most}
\APACinsertmetastar {%
wainer2007most}%
\begin{APACrefauthors}%
Wainer, H.%
\end{APACrefauthors}%
\unskip\
\newblock
\APACrefYearMonthDay{2007}{}{}.
\newblock
{\BBOQ}\APACrefatitle {The most dangerous equation} {The most dangerous
  equation}.{\BBCQ}
\newblock
\APACjournalVolNumPages{American Scientist}{95}{3}{249}.
\PrintBackRefs{\CurrentBib}

\bibitem [\protect \citeauthoryear {%
Wasserstein%
\ \BBA {} Lazar%
}{%
Wasserstein%
\ \BBA {} Lazar%
}{%
{\protect \APACyear {2016}}%
}]{%
wasserstein2016asa}
\APACinsertmetastar {%
wasserstein2016asa}%
\begin{APACrefauthors}%
Wasserstein, R\BPBI L.%
\BCBT {}\ \BBA {} Lazar, N\BPBI A.%
\end{APACrefauthors}%
\unskip\
\newblock
\APACrefYearMonthDay{2016}{}{}.
\newblock
\APACrefbtitle {The ASA statement on p-values: context, process, and purpose}
  {The asa statement on p-values: context, process, and purpose}\ (\BVOL~70)\
  (\BNUM~2).
\newblock
\APACaddressPublisher{}{Taylor \& Francis}.
\PrintBackRefs{\CurrentBib}

\bibitem [\protect \citeauthoryear {%
Weinberg%
}{%
Weinberg%
}{%
{\protect \APACyear {2022}}%
}]{%
Weinberg.2022}
\APACinsertmetastar {%
Weinberg.2022}%
\begin{APACrefauthors}%
Weinberg, L.%
\end{APACrefauthors}%
\unskip\
\newblock
\APACrefYearMonthDay{2022}{}{}.
\newblock
{\BBOQ}\APACrefatitle {Rethinking Fairness: An Interdisciplinary Survey of
  Critiques of Hegemonic ML Fairness Approaches} {Rethinking fairness: An
  interdisciplinary survey of critiques of hegemonic ml fairness
  approaches}.{\BBCQ}
\newblock
\APACjournalVolNumPages{Journal of Artificial Intelligence
  Research}{74}{}{75--109}.
\PrintBackRefs{\CurrentBib}

\bibitem [\protect \citeauthoryear {%
Westfall%
\ \BBA {} Young%
}{%
Westfall%
\ \BBA {} Young%
}{%
{\protect \APACyear {1993}}%
}]{%
westfall1993resampling}
\APACinsertmetastar {%
westfall1993resampling}%
\begin{APACrefauthors}%
Westfall, P\BPBI H.%
\BCBT {}\ \BBA {} Young, S\BPBI S.%
\end{APACrefauthors}%
\unskip\
\newblock
\APACrefYear{1993}.
\newblock
\APACrefbtitle {Resampling-based multiple testing: Examples and methods for
  p-value adjustment} {Resampling-based multiple testing: Examples and methods
  for p-value adjustment}\ (\BVOL~279).
\newblock
\APACaddressPublisher{}{John Wiley \& Sons}.
\PrintBackRefs{\CurrentBib}

\bibitem [\protect \citeauthoryear {%
White%
, van~der Ende%
\BCBL {}\ \BBA {} Nichols%
}{%
White%
\ \protect \BOthers {.}}{%
{\protect \APACyear {2019}}%
}]{%
white2019beyond}
\APACinsertmetastar {%
white2019beyond}%
\begin{APACrefauthors}%
White, T.%
, van~der Ende, J.%
\BCBL {}\ \BBA {} Nichols, T\BPBI E.%
\end{APACrefauthors}%
\unskip\
\newblock
\APACrefYearMonthDay{2019}{}{}.
\newblock
{\BBOQ}\APACrefatitle {Beyond Bonferroni revisited: concerns over inflated
  false positive research findings in the fields of conservation genetics,
  biology, and medicine} {Beyond bonferroni revisited: concerns over inflated
  false positive research findings in the fields of conservation genetics,
  biology, and medicine}.{\BBCQ}
\newblock
\APACjournalVolNumPages{Conservation Genetics}{20}{4}{927--937}.
\PrintBackRefs{\CurrentBib}

\bibitem [\protect \citeauthoryear {%
Wicherts%
\ \protect \BOthers {.}}{%
Wicherts%
\ \protect \BOthers {.}}{%
{\protect \APACyear {2016}}%
}]{%
wicherts2016degrees}
\APACinsertmetastar {%
wicherts2016degrees}%
\begin{APACrefauthors}%
Wicherts, J\BPBI M.%
, Veldkamp, C\BPBI L.%
, Augusteijn, H\BPBI E.%
, Bakker, M.%
, Van~Aert, R.%
\BCBL {}\ \BBA {} Van~Assen, M\BPBI A.%
\end{APACrefauthors}%
\unskip\
\newblock
\APACrefYearMonthDay{2016}{}{}.
\newblock
{\BBOQ}\APACrefatitle {Degrees of freedom in planning, running, analyzing, and
  reporting psychological studies: A checklist to avoid p-hacking} {Degrees of
  freedom in planning, running, analyzing, and reporting psychological studies:
  A checklist to avoid p-hacking}.{\BBCQ}
\newblock
\APACjournalVolNumPages{Frontiers in psychology}{}{}{1832}.
\PrintBackRefs{\CurrentBib}

\bibitem [\protect \citeauthoryear {%
Xivuri%
\ \BBA {} Twinomurinzi%
}{%
Xivuri%
\ \BBA {} Twinomurinzi%
}{%
{\protect \APACyear {2021}}%
}]{%
Xivuri.2021}
\APACinsertmetastar {%
Xivuri.2021}%
\begin{APACrefauthors}%
Xivuri, K.%
\BCBT {}\ \BBA {} Twinomurinzi, H.%
\end{APACrefauthors}%
\unskip\
\newblock
\APACrefYearMonthDay{2021}{}{}.
\newblock
{\BBOQ}\APACrefatitle {A Systematic Review of Fairness in Artificial
  Intelligence Algorithms} {A systematic review of fairness in artificial
  intelligence algorithms}.{\BBCQ}
\newblock
\BIn{} D.~Dennehy, A.~Griva, N.~Pouloudi, Y\BPBI K.~Dwivedi, I.~Pappas\BCBL {}\
  \BBA {} M.~M{\"a}ntym{\"a}ki\ (\BEDS), \APACrefbtitle {Responsible AI and
  Analytics for an Ethical and Inclusive Digitized Society} {Responsible ai and
  analytics for an ethical and inclusive digitized society}\ (\BPGS\ 271--284).
\newblock
\APACaddressPublisher{Cham}{{Springer International Publishing}}.
\PrintBackRefs{\CurrentBib}

\bibitem [\protect \citeauthoryear {%
Zemel%
, Wu%
, Swersky%
, Pitassi%
\BCBL {}\ \BBA {} Dwork%
}{%
Zemel%
\ \protect \BOthers {.}}{%
{\protect \APACyear {2013}}%
}]{%
zemel2013learning}
\APACinsertmetastar {%
zemel2013learning}%
\begin{APACrefauthors}%
Zemel, R.%
, Wu, Y.%
, Swersky, K.%
, Pitassi, T.%
\BCBL {}\ \BBA {} Dwork, C.%
\end{APACrefauthors}%
\unskip\
\newblock
\APACrefYearMonthDay{2013}{}{}.
\newblock
{\BBOQ}\APACrefatitle {Learning fair representations} {Learning fair
  representations}.{\BBCQ}
\newblock
\BIn{} \APACrefbtitle {International conference on machine learning}
  {International conference on machine learning}\ (\BPGS\ 325--333).
\PrintBackRefs{\CurrentBib}

\bibitem [\protect \citeauthoryear {%
{\v{Z}}liobait{\.e}%
}{%
{\v{Z}}liobait{\.e}%
}{%
{\protect \APACyear {2017}}%
}]{%
vzliobaite2017measuring}
\APACinsertmetastar {%
vzliobaite2017measuring}%
\begin{APACrefauthors}%
{\v{Z}}liobait{\.e}, I.%
\end{APACrefauthors}%
\unskip\
\newblock
\APACrefYearMonthDay{2017}{}{}.
\newblock
{\BBOQ}\APACrefatitle {Measuring discrimination in algorithmic decision making}
  {Measuring discrimination in algorithmic decision making}.{\BBCQ}
\newblock
\APACjournalVolNumPages{Data Mining and Knowledge
  Discovery}{31}{4}{1060--1089}.
\PrintBackRefs{\CurrentBib}

\end{thebibliography}
\clearpage

\appendix
\section{Appendix}
\subsection{Figure 1 with equal opportunity and statistical parity }

\begin{figure}[h!]
	\includegraphics[width=0.99\textwidth]{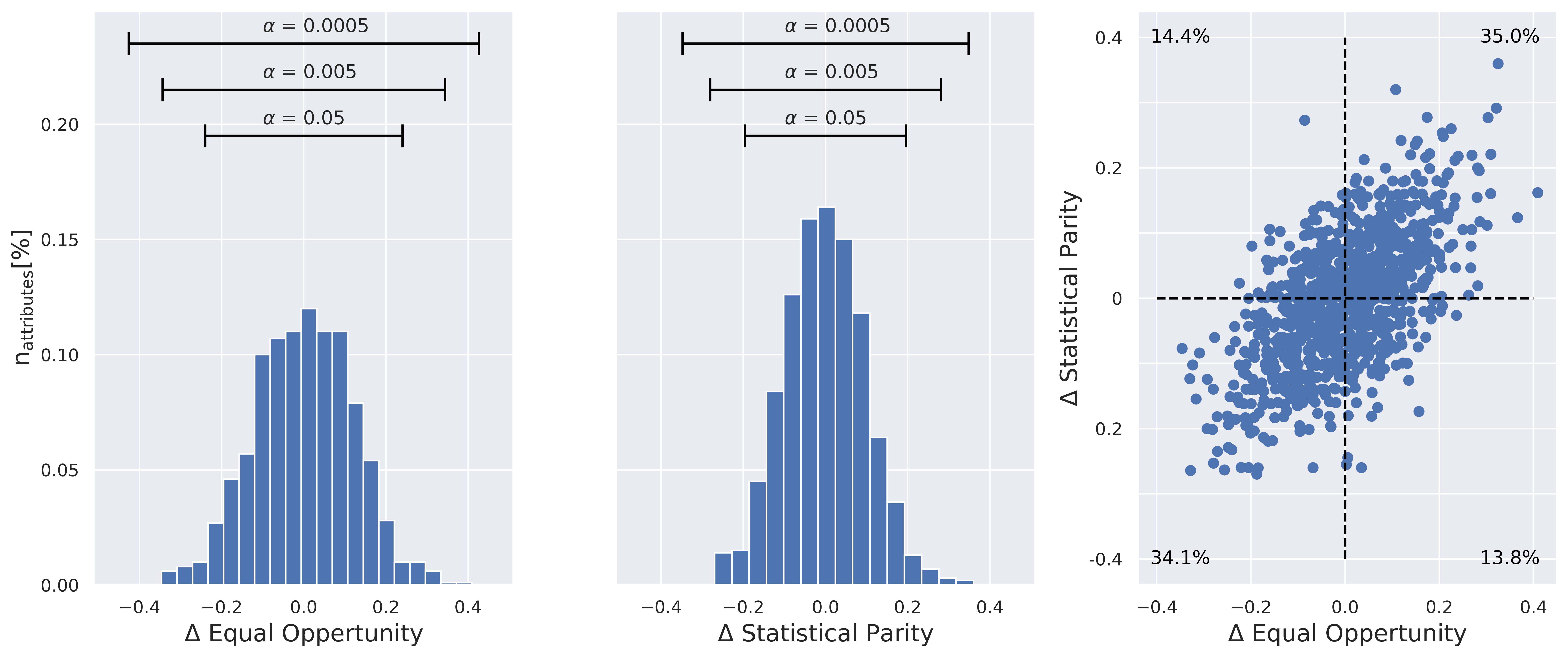}
	\caption{Intra-metric fairness hacking for equal opportunity and statistical parity. Plot conventions as in Figure \ref{fig::Figure11}.}
	\label{fig::Figure12}
\end{figure}
\clearpage
\subsection{Equal opportunity and statistical parity in the wild}
     \begin{figure}[h]
         \centering
         \includegraphics[width=1\textwidth]{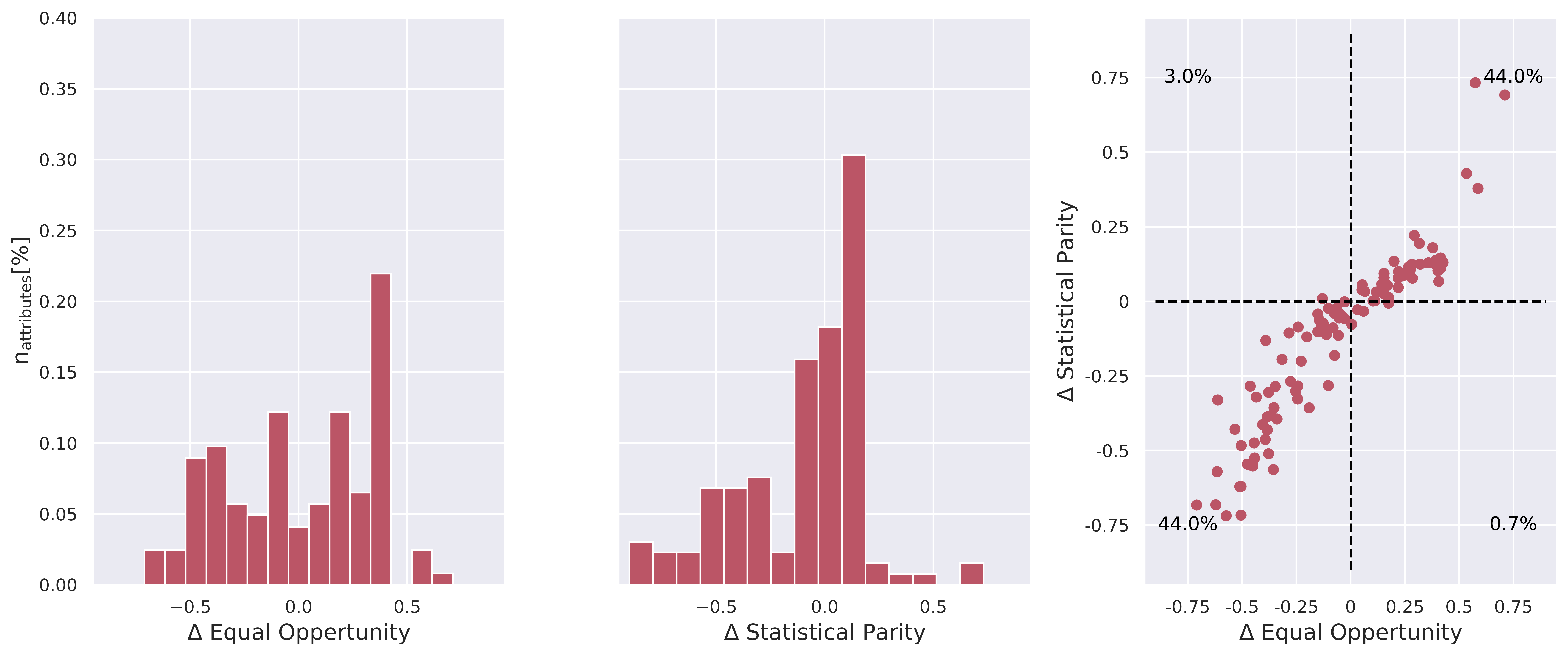}
         \caption{Equal opportunity and statistical parity for all 134 binary attributes. Plot conventions as in Figure \ref{fig::Figure11}. The number of items belonging to the protected or unprotected group differs between actual attributes in the MEPS dataset (race, sex, etc.). Thus, we cannot calculate a single null hypothesis confidence interval.}
         \label{fig:4_1}
     \end{figure}
\clearpage
\subsection{Individual fairness metrics}

\begin{figure}[h!]
	\includegraphics[width=0.8\textwidth]{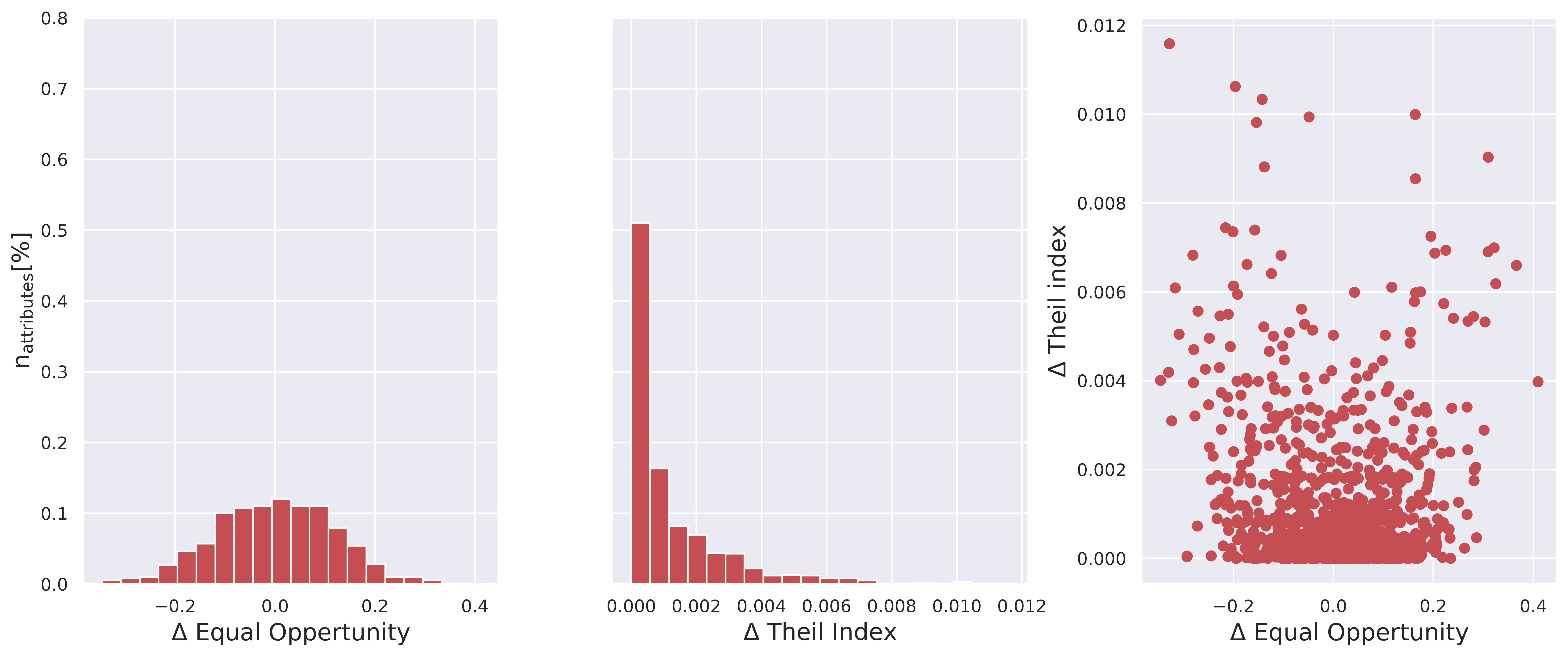}
	\caption{Equal opportunity and the Theil index for our hypothetical data. Plot conventions as in Figure \ref{fig::Figure11}.}
	\label{fig::FigureTheil}
\end{figure}

\begin{figure}[h!]
	\includegraphics[width=0.8\textwidth]{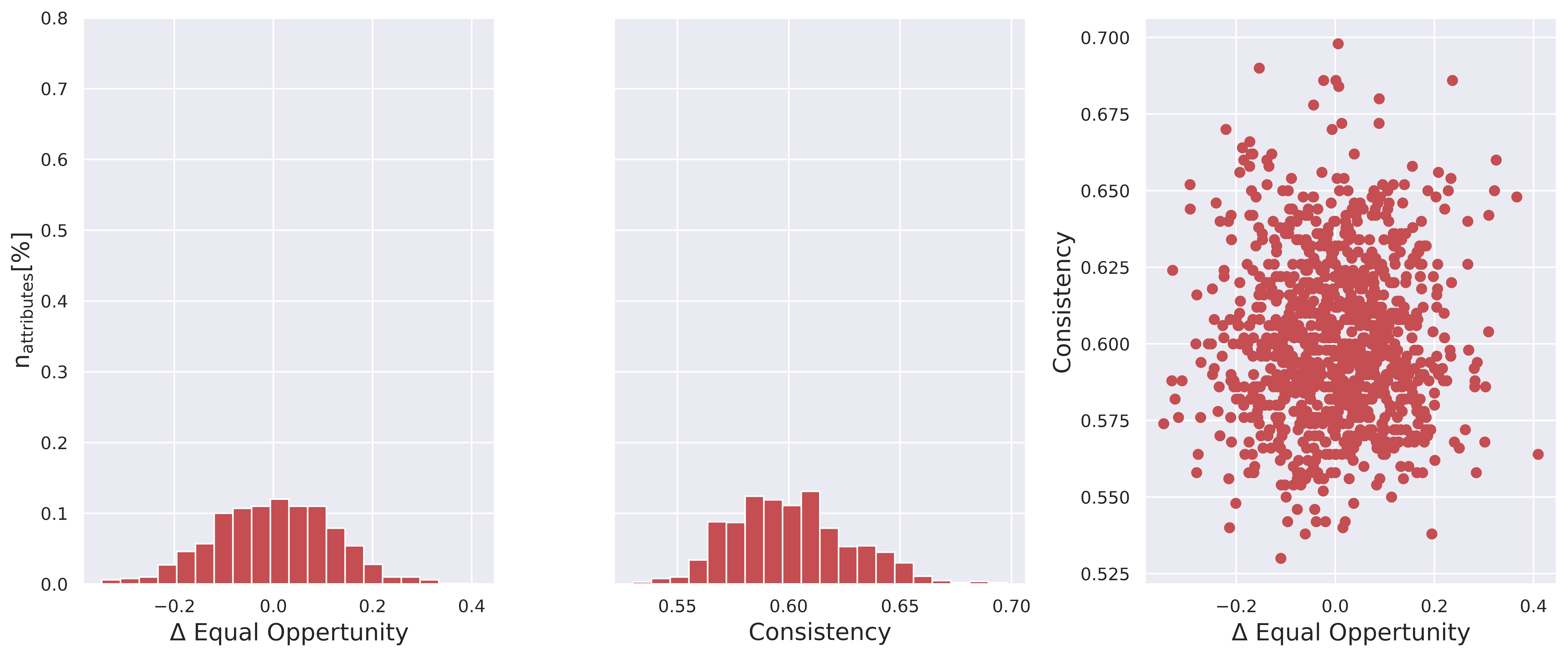}
	\caption{Equal opportunity and consistency for our hypothetical data. Plot conventions as in Figure \ref{fig::Figure11}.}
	\label{fig::Consistency}
\end{figure}
\clearpage
\subsection{Inter-metric fairness hacking for the Folktables dataset}

\begin{figure}[h!]
	\includegraphics[width=0.6\textwidth]{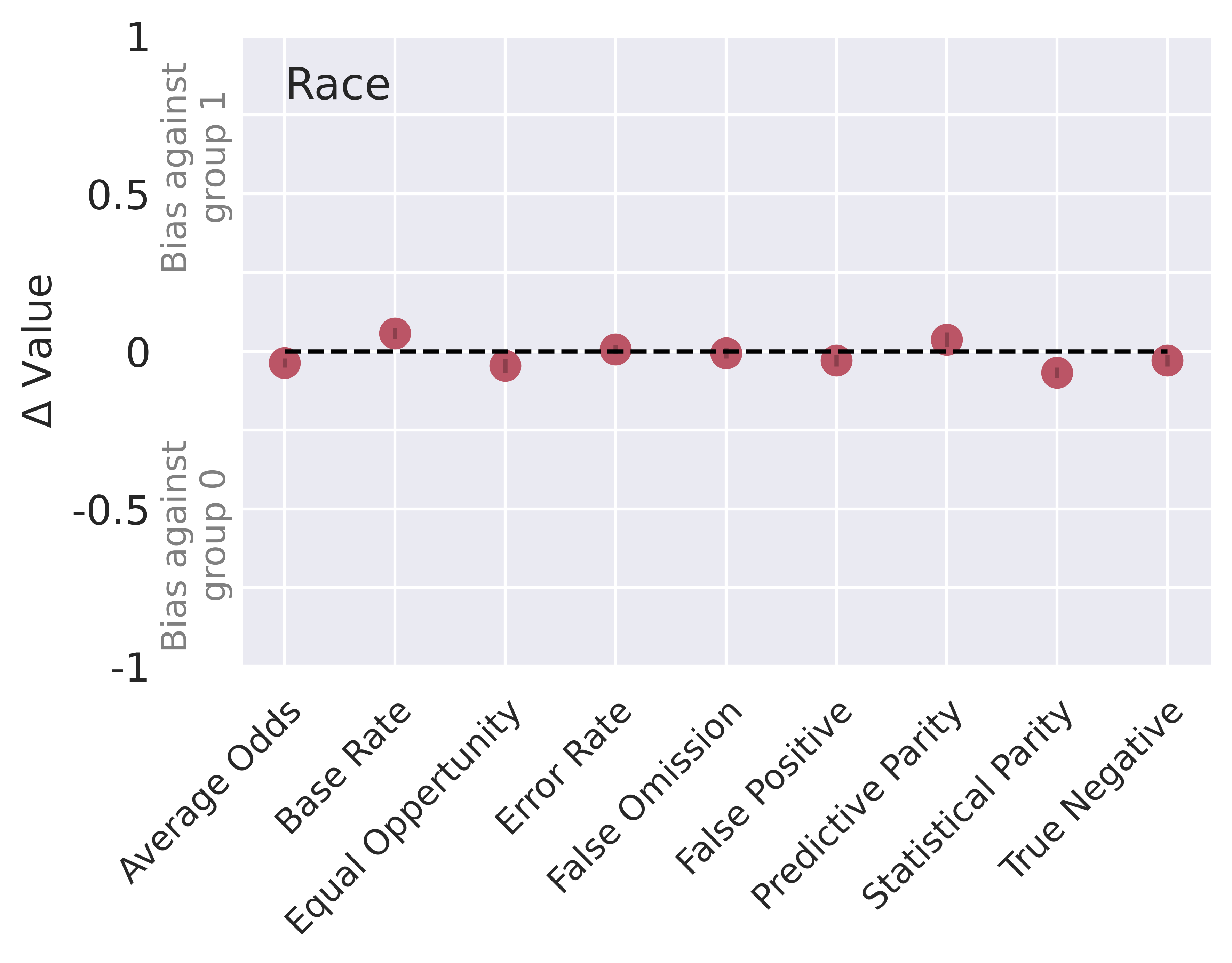}
	\caption{Data from the Folktables dataset. The provided logistic regression example model had an accuracy of 77.2\%. Only the race attribute is used. Plot conventions as in Figure \ref{fig:3_ab}.}
	\label{fig::FigureFolkTables}
\end{figure}

\end{document}